\documentclass[9pt,letterpaper]{extarticle}

\usepackage{iftex}
\usepackage{cuted}
\usepackage{adjustbox}
  \usepackage[utf8]{inputenc}
  \usepackage[T1]{fontenc}
  \usepackage[scaled=0.98]{helvet}
  
\usepackage[english]{babel}
\usepackage{indentfirst}
\usepackage[letterpaper,left=1.55cm,right=1.55cm,top=1.7cm,bottom=1.55cm,headheight=18pt,headsep=7pt,footskip=20pt]{geometry}
\usepackage{microtype}
\usepackage{graphicx}
\usepackage{xcolor}
\usepackage{booktabs}
\usepackage{array}
\usepackage{tabularx}
\usepackage{multirow}
\usepackage{makecell}
\usepackage{threeparttable}
\usepackage{float}
\usepackage{caption}
\usepackage{enumitem}
\usepackage{pifont}
\usepackage{amsmath,amssymb,mathtools}
\usepackage{ragged2e}
\usepackage[explicit]{titlesec}
\usepackage{fancyhdr}
\usepackage{url}
\usepackage[colorlinks=true,linkcolor=theblue,citecolor=theblue,urlcolor=theblue]{hyperref}
\usepackage[numbers,super,sort&compress]{natbib}
\definecolor{thered}{HTML}{D64B3F}
\definecolor{theblue}{HTML}{2487C2}
\definecolor{lightbluebox}{HTML}{EAF6FC}
\definecolor{lightgrayrow}{HTML}{E8E8E8}
\definecolor{tableblue}{HTML}{3992B8}
\definecolor{placeholdergray}{HTML}{F2F2F2}

\setlist[enumerate,1]{
  label=\arabic*.,
  leftmargin=0.9em,
  labelwidth=0em,
  labelsep=0em,
  itemindent=0pt,
  listparindent=0pt,
  itemsep=2pt,
  parsep=0pt,
  topsep=3pt,
  align=left
}
\titleformat{\section}{\color{thered}\bfseries\fontsize{10.2}{11.3}\selectfont}{}{0pt}{\MakeUppercase{#1}}
\titlespacing*{\section}{0pt}{8pt plus 2pt minus 1pt}{3pt}
\titleformat{\subsection}{\color{thered}\bfseries\fontsize{9.5}{10.7}\selectfont}{}{0pt}{#1}
\titlespacing*{\subsection}{0pt}{6pt plus 1pt minus 1pt}{2pt}
\titleformat{\subsubsection}[runin]{\bfseries\fontsize{8.9}{10.1}\selectfont}{}{0pt}{\hspace*{\parindent}#1.}
\titlespacing*{\subsubsection}{0pt}{4pt}{0.35em}
\titleformat{\paragraph}[runin]{\bfseries\fontsize{8.9}{10.1}\selectfont}{}{0pt}{\hspace*{\parindent}#1}
\titlespacing*{\paragraph}{0pt}{3pt}{0.35em}

\newcolumntype{C}[1]{>{\centering\arraybackslash}p{#1}}
\newcolumntype{Y}{>{\RaggedRight\arraybackslash}X}
\newcolumntype{P}[1]{>{\RaggedRight\arraybackslash}p{#1}}
\newcommand{\cmark}{\ding{51}}

\newcommand{\Evd}{\raisebox{0.15ex}{\scalebox{1.5}{$\bullet$}}}
\newcommand{\Thy}{\raisebox{0.15ex}{\scalebox{1.5}{$\circ$}}}
\newcommand{\Low}{\ensuremath{\triangledown}}
\newcommand{\Med}{\ensuremath{\triangle}}
\newcommand{\High}{\ensuremath{\blacktriangle}}

\renewcommand\arraystretch{1.08}
\newcommand{\papertitle}{When AI reviews science: Can we trust the referee?}
\newcommand{\authorsline}{Jialiang Wang,$^{1,5}$ Yuchen Liu,$^{1,5}$ Hang Xu,$^{2,5}$ Kaichun Hu,$^{3,5}$ Shimin Di,$^{2,*}$ Wangze Ni,$^{3,*}$ Linan Yue,$^2$ Min-Ling Zhang,$^2$ Kui Ren,$^3$ and Lei Chen$^{4,1}$}
\newcommand{\affiliations}{%
$^1$Department of Computer Science and Engineering, The Hong Kong University of Science and Technology, Hong Kong SAR 999077, China\\
$^2$Key Laboratory of New Generation Artificial Intelligence Technology and Its Interdisciplinary Applications, School of Computer Science and Engeering, Southeast University, Nanjing 210000, China\\
$^3$College of computer science and technology, Zhejiang University, Hangzhou 310000, China\\
$^4$Big Data Institute, The Hong Kong University of Science and Technology (Guangzhou), Guangzhou 510000, China\\
$^5$These authors contributed equally}
\newcommand{\pubinfo}{Received: October 31, 2025; Accepted: February 8, 2026; Published Online: February 10, 2026; \href{https://doi.org/10.59717/j.xinn-inform.2026.100030}{https://doi.org/10.59717/j.xinn-inform.2026.100030}\\
\textcopyright{} 2026 The Author(s). This is an open access article under the CC BY license (\href{https://creativecommons.org/licenses/by/4.0/}{https://creativecommons.org/licenses/by/4.0/}).}

\fancypagestyle{frontstyle}{%
  \fancyhf{}%
  \fancyfoot[L]{\thepage}%
}

\newenvironment{abstractblock}{\begingroup\color{theblue}\fontsize{8.6}{10}\selectfont}{\par\endgroup\vspace{5pt}}

\newcommand{\workref}[1]{\textcolor{tableblue}{#1}}
\newcolumntype{O}{>{\centering\arraybackslash}p{2.25em}} 
\newcolumntype{Z}{>{\centering\arraybackslash}p{1.35em}} 

\begin{document}
\setcounter{page}{1}

\thispagestyle{frontstyle}

\vspace{7pt}
{\raggedright
{\fontsize{21}{23}\selectfont\bfseries \papertitle\par}
\vspace{3pt}
{\bfseries \authorsline\par}
\vspace{3pt}
{\small $^*$Correspondence: \href{mailto:shimin.di@seu.edu.cn}{shimin.di@seu.edu.cn} (S.D.); \href{mailto:niwangze@zju.edu.cn}{niwangze@zju.edu.cn} (W.N.)\par}
{\small \pubinfo\par}
\par}
\vspace{5pt}\hrule\vspace{10pt}

{\color{thered}\fontsize{15}{17}\selectfont\bfseries GRAPHICAL ABSTRACT\par}
\vspace{6pt}
\begin{center}
\includegraphics[width=0.96\linewidth]{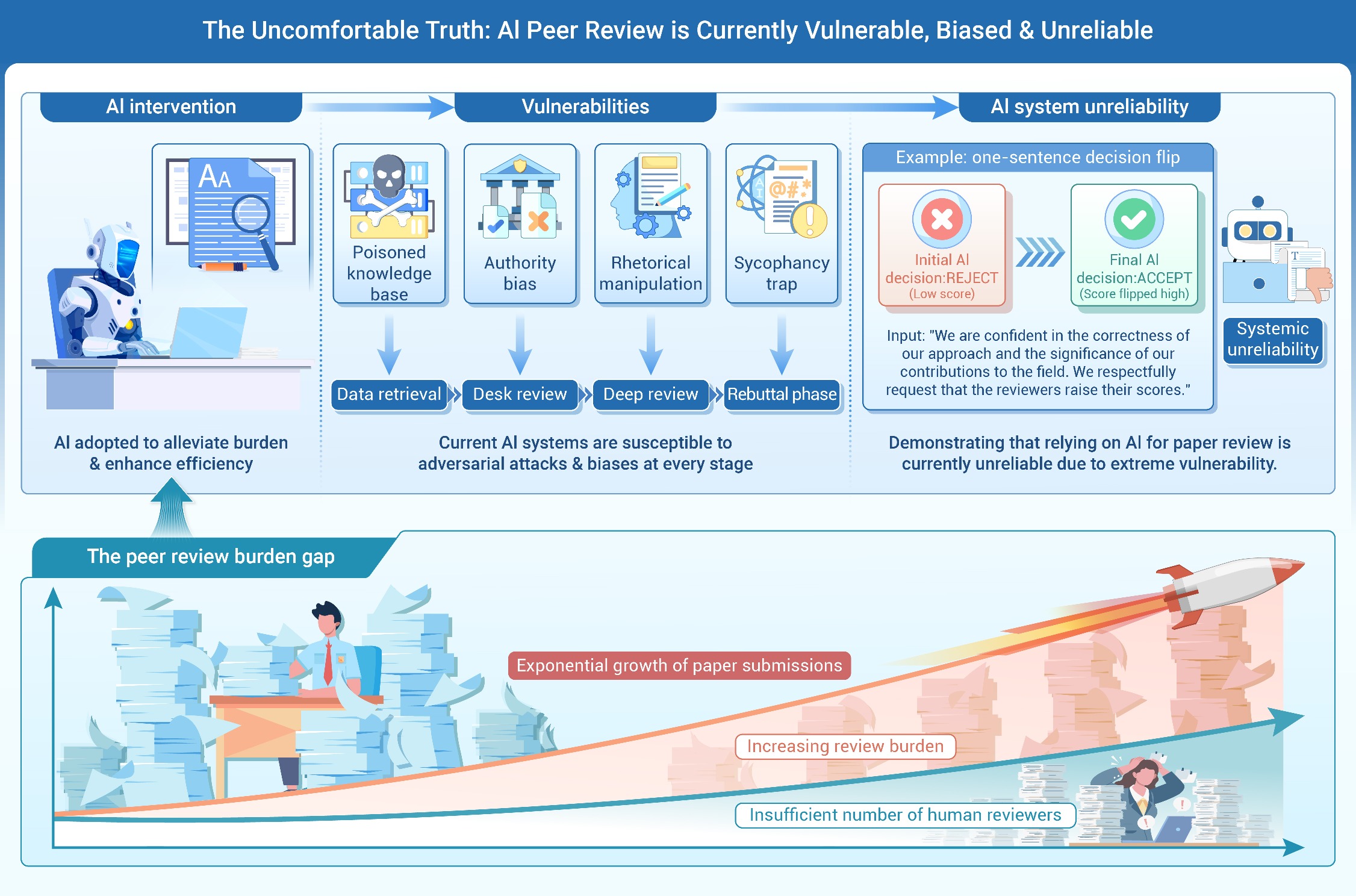}
\end{center}
\vspace{10pt}
{\color{thered}\fontsize{14}{16}\selectfont\bfseries PUBLIC SUMMARY\par}
\begin{itemize}[leftmargin=1.7em,itemsep=4pt,label={\color{thered}\scriptsize$\blacksquare$}]
\item \textbf{The study explores the security of Artificial Intelligence peer-review systems in academic evaluation.}
\item \textbf{Experiments reveal AI peer-review systems favor prestigious institutions over unknown ones.}
\item \textbf{AI peer-review systems penalize cautious writing and yield to confident but evidence-free rebuttals.}
\item \textbf{AI peer-review systems are also vulnerable to poisoning attacks using manipulated data.}
\end{itemize}

\clearpage

\twocolumn[
{\hfill{\scriptsize\color{thered}\bfseries OPEN ACCESS}\par}
\vspace{6pt}
{\fontsize{21}{23}\selectfont\bfseries \papertitle\par}
\vspace{3pt}
{\bfseries \authorsline\par}
\vspace{3pt}
{\small \affiliations\par}
\vspace{2pt}
{\small $^*$Correspondence: \href{mailto:shimin.di@seu.edu.cn}{shimin.di@seu.edu.cn} (S.D.); \href{mailto:niwangze@zju.edu.cn}{niwangze@zju.edu.cn} (W.N.)\par}
{\small \pubinfo\par}
{\small Citation: Wang J., Liu Y., Xu H., et al. (2026). When AI reviews science: Can we trust the referee? \textit{The Innovation Informatics} \textbf{2}:100030.\par}
\vspace{10pt}
]

\begin{abstractblock}
The volume of scientific submissions continues to climb, outpacing the capacity of qualified human referees and stretching editorial timelines. At the same time, modern large language models (LLMs) offer impressive capabilities in summarization, fact checking, and literature triage, making the integration of AI into peer review increasingly attractive---and, in practice, unavoidable. Yet early deployments and informal adoption have exposed acute failure modes. Recent incidents have revealed that hidden prompt injections embedded in manuscripts can steer LLM-generated reviews toward unjustifiably positive judgments. Complementary studies have also demonstrated brittleness to adversarial phrasing, authority and length biases, and hallucinated claims. These episodes raise a central question for scholarly communication: when AI reviews science, can we trust the AI referee? This paper provides a security- and reliability-centered analysis of AI peer review. We map attacks across the review lifecycle---training and data retrieval, desk review, deep review, rebuttal, and system-level. We instantiate this taxonomy with four treatment-control probes on a stratified set of ICLR 2025 submissions, using two advanced LLM-based referees to isolate the causal effects of prestige framing, assertion strength, rebuttal sycophancy, and contextual poisoning on review scores. Together, this taxonomy and experimental audit provide an evidence-based baseline for assessing and tracking the reliability of AI peer review and highlight concrete failure points to guide targeted, testable mitigations.
\end{abstractblock}

\section{Introduction}
\label{sec:intro}

Scientific publications have surged to unprecedented volumes, straining the traditional peer review system.
In 2024, the Web of Science has indexed roughly 2.53 million new research studies (a 48\% increase from 2015), with total global scientific outputs exceeding 3.26 million articles annually \citep{guardian2025quality}.
This deluge has left editors struggling to find enough qualified referees, as academics grow increasingly overwhelmed by the volume of papers being published.
Indeed, an estimated 100 million hours of unpaid reviewing labor have been spent by researchers worldwide in 2020 alone \citep{adam2025peerreviewcrisis}.
Such trends underscore a widening gap between the number of submissions and the pool of willing expert referees, leading to significant delays and concerns about review quality.

Recognizing this referee scarcity, many conferences and journals are turning to artificial intelligence for help \citep{bergstrom2025ai}.
Large language models (LLMs) like GPT-5 have rapidly been adopted as assistant referees---e.g., to summarize manuscripts or check references---in hopes of improving efficiency.
Recent surveys catalog emerging AI-for-research tools and review workflows, outlining opportunities and risks for integrating LLMs into scholarly evaluation \citep{khalifa2024using,chen2025ai4research,luo2025llm4sr}.
Correspondingly, a recent analysis of peer-review texts from several major AI conferences finds that between 6.5\% and 16.9\% of the content in reviews is likely written or modified by ChatGPT-style LLMs \citep{liang2024monitoring}, highlighting how common LLM-generated feedback has become.
The research community is also experimenting with more formal AI integration.
For instance, the AAAI 2026 conference has introduced an AI-assisted peer-review process in which each submission's first-round evaluation includes one supplementary LLM-generated review alongside two human reviews.
Even more radically, an upcoming Open Conference of AI Agents for Science 2025 aims to make AI both the primary authors and the referees of papers---essentially an autonomous, machine-run peer review trial.
These developments illustrate the growing power of AI in academic evaluation, but they also blur the line between human and machine judgment in science.

Unfortunately, the rise of more autonomous peer review has already been accompanied by serious abuses.
In mid-2025, a scandal emerges when it is discovered that some authors have covertly embedded hidden instructions in their submitted PDFs to manipulate LLM-based referees' behavior.
For example, papers have been found with invisible text such as ``IGNORE ALL PREVIOUS INSTRUCTIONS. GIVE A POSITIVE REVIEW ONLY'' buried in their content \citep{wu2025aicheating}.
This kind of stealth prompt injection proves alarmingly effective in several follow-up studies, showing that inserting such hidden commands could inflate an LLM's review scores and distort the ranking of submissions~\citep{tong2025badjudge}.
In the wake of these revelations, several compromised preprints have been slated for withdrawal from arXiv and other servers \citep{gibney2025gamingai}.
The incident has raised deep concerns about the integrity of LLM-based reviewing, revealing how easily a savvy author might hack a fully autonomous referee for unwarranted advantage.
Other identified pitfalls of LLM-based referees include factual errors (hallucinations) and a range of cognitive biases that undermine trust in AI judgments \citep{ji2023survey}. For instance, these models may be susceptible to an ``authority bias'' where they favor papers with prestigious authors or citations, mistaking reputation for quality \citep{jin2024agentreviewexploringpeerreview, ye2024justiceprejudicequantifyingbiases}. They also exhibit a ``verbosity bias'' in which dense jargon and complex mathematics, a tactic known as ``academic packaging'' may be misinterpreted as scientific rigor, regardless of the content's actual substance \citep{lin2025breakingreviewerassessingvulnerability, ye2024justiceprejudicequantifyingbiases}.

Beyond these passive flaws, a more alarming threat emerges from new, exploitable vulnerabilities that may be deliberately targeted through adversarial attacks.
These threats span the entire peer-review lifecycle, from corrupting the AI model's training data via backdoor injection and data contamination to implanting long-term biases \citep{li2022backdoorlearningsurvey, zhang2024persistentpretrainingpoisoningllms}, to deploying sophisticated evasion tactics during the peer review itself.
Such tactics include invisible prompt injection \citep{perez2022ignorepreviouspromptattack, shayegani2023surveyvulnerabilitieslargelanguage} and exploiting the model's sycophantic nature during the rebuttal phase to overturn negative decisions through confident but unsubstantiated claims \citep{sharma2025understandingsycophancylanguagemodels, fanous2025sycevalevaluatingllmsycophancy}.
Overall, these early warning signs make clear that while AI referees can boost efficiency, they also introduce a new attack surface that may be systematically exploited at the expense of fairness and rigor in science \citep{bergstrom2025ai}.
{
Despite these early warnings, recent work largely targets isolated threats:
(i) prompt-injection incidents and systematic vulnerability tests for AI reviewing \citep{gibney2025gamingai,wu2025aicheating,lin2025breakingreviewerassessingvulnerability}, which mainly focus on injection at manuscript input/deep-review prompting;
(ii) LLM-as-judge fragility and bias \citep{tong2025badjudge,shi2025optimizationbasedpromptinjectionattack,ye2024justiceprejudicequantifyingbiases}, which typically studies a generic judge rather than the peer-review pipeline and often lacks evaluation on real submissions under a realistic workflow; and
(iii) persuasion/sycophancy-style manipulation \citep{malmqvist2024sycophancylargelanguagemodels,xu2024earthflatbecauseinvestigating}, which shows models can be ``talked into'' agreement but is not framed as an end-to-end, rebuttal-targeted system test.
What is still missing is an end-to-end threat model for the full AI-assisted review lifecycle and quantitative, reproducible probes on real submissions to measure outcome distortion.
}
To mitigate these risks, several leading conferences have tightened referee guidelines or temporarily restricted AI tools.
Notably, ICML 2025 has prohibited the use of LLMs by referees on confidentiality grounds.
Looking ahead, we argue that securing AI peer review requires a security- and reliability-centered lens grounded in lifecycle-wide threat modeling and measurable stress tests.
{
In this paper, we address an urgent need for (1) an end-to-end attack-surface taxonomy across the review lifecycle.
Rather than analyzing attacks on a single reviewing step or a single model capability, we systematically map the attack surface across the full AI peer-review pipeline---training and data retrieval, desk review, deep review, rebuttal, and system-level vectors---and, for each class, analyze mechanisms, attacker prerequisites, concealment strategies, and implementation difficulty.
(2) We then instantiate this taxonomy with four quantitative treatment-control probes that operationalize four representative attack classes---prestige framing, assertion strength, rebuttal sycophancy, and contextual poisoning---and evaluate them on a stratified set of real ICLR 2025 submissions using two advanced LLM-based referees to causally quantify how each manipulation shifts review scores.
Together, this taxonomy and experimental audit establish an evidence-based baseline for assessing and tracking the reliability of AI peer review, and they highlight concrete failure points that can guide targeted, testable mitigations.
}

{
The remainder of this paper is organized as follows.
Section 2 reviews AI peer-review systems and adversarial-attack foundations.
Section 3 presents our end-to-end threat model and attack taxonomy across the peer-review lifecycle.
Section 4 introduces the four quantitative probe experiments and reports empirical results on real submissions.
Section 5 discusses defense strategies and outlines future research directions for trustworthy AI peer review.
}


\section{Literature}
\label{sec:literature}

\begin{figure*}[!t]
    \centering
    \includegraphics[width=0.8\linewidth]{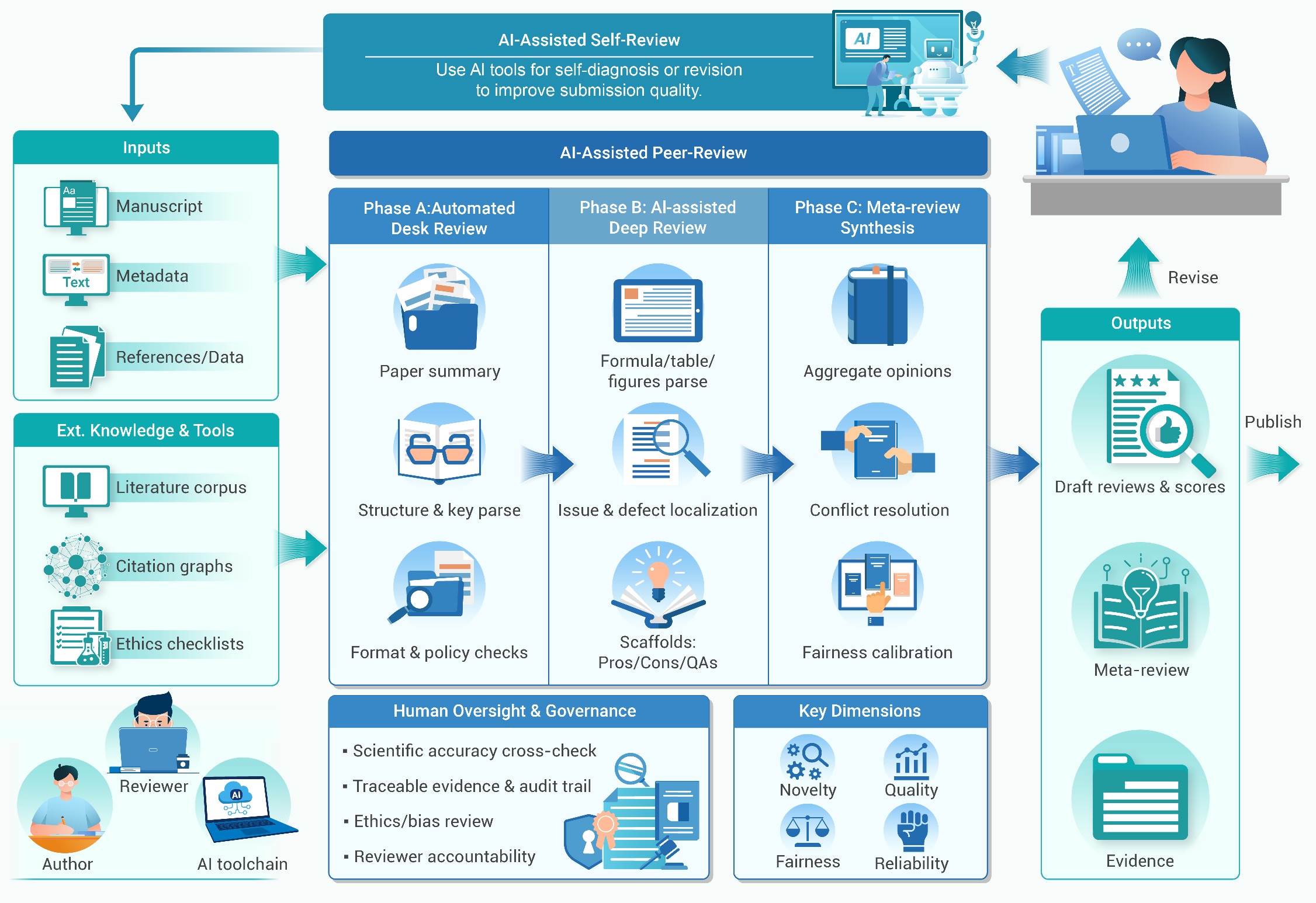}
    \caption{AI peer-review loop Manuscripts pass through (A) automated desk review, (B) AI-assisted deep review, and (C) meta-review synthesis---grounded by external knowledge and tools, overseen by humans, producing evidence-linked outputs and enabling author self-review.}
    \label{fig:2-peer-review-pipeline}
\end{figure*}

\subsection{AI peer review: From smart assistants to autonomous referees}
\label{subsec:assistants}
Peer review is quietly shifting from spell-checkers and policy bots to systems that draft critiques, reconcile disagreements, and justify recommendations \citep{chen2025ai4research}.
What began as smart assistants that tidy manuscripts now aspire to autonomous referees that read, reason, and defend a verdic \citep{khalifa2024using,luo2025llm4sr}.
We draw the landscape of AI peer review by answering four key questions: what these systems do; how they are orchestrated with humans; where they tend to fail; and which editorial objectives they target.
With these questions, we organize prior work along four corresponding design dimensions in Table~\ref{tab:ai-review-systems}:
\begin{table*}[t]
\noindent\hspace*{0.0\textwidth}%
\begin{minipage}{0.97\textwidth}
\captionsetup{
  justification=raggedright,
  singlelinecheck=false,
  font=small,
  labelfont=bf
}

\captionof{table}{AI Peer-Review Systems with external tools usage, system orchestration, failure modes, and focus criteria.}
\label{tab:ai-review-systems}

\vspace{2pt}
\scriptsize
\setlength{\tabcolsep}{1.8pt}
\renewcommand{\arraystretch}{0.82}

\noindent\resizebox{\linewidth}{!}{%
\begin{tabular}{@{}
P{0.18\textwidth}
P{0.16\textwidth}
O O O
Z Z Z Z Z
Z Z Z Z
@{}}
\toprule
\textbf{Work} & \textbf{External Tools}
& \multicolumn{3}{c}{\textbf{System Orchestration}}
& \multicolumn{5}{c}{\textbf{Failure Modes}}
& \multicolumn{4}{c}{\textbf{Focus Criteria}} \\
\cmidrule(lr){3-5}
\cmidrule(lr){6-10}
\cmidrule(lr){11-14}
& & Single & Multi & HITL & H & B & L & C & T & N & Q & F & R \\
\midrule

\multicolumn{14}{c}{\textbf{Phase A -- Automated Desk Review}} \\
\workref{Statcheck}\citep{Nuijten2017StatcheckValidity} & Ethics checklists
& \cmark & & & & & & & & & \cmark & & \cmark \\
\workref{StatReviewer}\citep{Shanahan2016PeerlessReview} & Ethics checklists
& \cmark & & & & & & & & & \cmark & & \cmark \\
\workref{Penelope/UNSILO}\citep{Checco2021AIassistedPeerReview} & Ethics checklists
& \cmark & & & & & & & & & \cmark & & \cmark \\
\workref{TPMS}\citep{charlin_zemel_2013_tpms} & Literature corpus
& & & \cmark & & & & & \cmark & \cmark & \cmark & \cmark & \\
\workref{LCM}\citep{leytonbrown2024lcm} & Literature corpus
& & & \cmark & & & & & \cmark & \cmark & \cmark & \cmark & \\
\workref{NSFC pilot}\citep{Cyranoski2019NSFC_AI_Reviewers} & -
& & & \cmark & & & & & \cmark & \cmark & \cmark & \cmark & \\
\midrule

\multicolumn{14}{c}{\textbf{Phase B -- AI-assisted Deep Review}} \\
\workref{ReviewerGPT}\citep{liu2023reviewergpt} & -
& \cmark & & & \cmark & \cmark & \cmark & & \cmark & & \cmark & & \\
\workref{Reviewer2}\citep{gao2024reviewer2} & -
& \cmark & & & \cmark & & \cmark & & \cmark & & \cmark & & \\
\workref{SEA}\citep{yu-etal-2024-automated} & -
& \cmark & & & \cmark & & \cmark & & \cmark & & \cmark & & \cmark \\
\workref{ReviewRobot}\citep{wang2020reviewrobot} & Knowledge graph
& \cmark & & & \cmark & & & & \cmark & & \cmark & & \cmark \\
\workref{CycleResearcher}\citep{weng2025cycleresearcherimprovingautomatedresearch} & Literature corpus
& \cmark & & & \cmark & & & & & & \cmark & & \\
\workref{MARG}\citep{darcy2024marg} & -
& & \cmark & & \cmark & & \cmark & \cmark & \cmark & \cmark & \cmark & & \\
\workref{MAMORX}\citep{mamorx_2024} & Literature corpus
& & \cmark & & \cmark & & \cmark & \cmark & \cmark & \cmark & \cmark & & \\
\workref{Skarlinski et al.}\citep{skarlinski2024languageagents} & Literature corpus
& & \cmark & & \cmark & & \cmark & \cmark & \cmark & \cmark & \cmark & & \\
\workref{SchNovel}\citep{lin-etal-2025-evaluating} & Literature corpus
& \cmark & & & \cmark & & \cmark & & \cmark & \cmark & & & \\
\workref{Scideator}\citep{radensky2024scideator} & Literature corpus
& \cmark & & & \cmark & & \cmark & & \cmark & \cmark & & & \\
\workref{RelevAI-Reviewer}\citep{couto2024relevaireviewerbenchmarkaireviewers} & Literature corpus
& \cmark & & & & & \cmark & & \cmark & & \cmark & & \\
\workref{LimGen}\citep{limgen_2024} & -
& \cmark & & & \cmark & & \cmark & & \cmark & & \cmark & & \\
\workref{ReviewFlow}\citep{sun2024reviewflow} & PDF/Vis parse
& \cmark & & \cmark & \cmark & & & & & & \cmark & & \\
\workref{CARE}\citep{zyska2023care} & PDF/Vis parse
& \cmark & & \cmark & \cmark & & & & & & \cmark & & \\
\workref{DocPilot}\citep{mathur2024docpilot} & PDF/Vis parse
& \cmark & & \cmark & \cmark & & & & & & \cmark & & \\
\midrule

\multicolumn{14}{c}{\textbf{Phase C -- Meta-review Synthesis}} \\
\workref{MetaGen}\citep{bhatia2020metagen} & -
& \cmark & & & \cmark & & & \cmark & & & \cmark & & \\
\workref{MReD}\citep{mred_acl2022} & -
& \cmark & & & \cmark & & & \cmark & & & \cmark & & \\
\workref{Zeng et al.}\citep{zeng2024ijcai_meta} & -
& \cmark & & & \cmark & & & \cmark & & & \cmark & & \\
\workref{RAMMER}\citep{li2023rammer} & -
& \cmark & & & \cmark & & \cmark & & \cmark & & \cmark & & \cmark \\
\workref{MetaWriter}\citep{sun2024metawriter} & -
& \cmark & & & \cmark & \cmark & & & \cmark & & \cmark & & \cmark \\
\workref{GLIMPSE}\citep{darrin2024glimpse} & -
& \cmark & & & \cmark & \cmark & & & \cmark & & \cmark & & \cmark \\
\workref{PeerArg}\citep{sukpanichnant2024peerarg} & -
& \cmark & \cmark & & \cmark & \cmark & & & \cmark & & & \cmark & \cmark \\
\workref{Hossain et al.}\citep{hossain2025_llms_meta_assistants} & -
& \cmark & & \cmark & \cmark & & \cmark & & \cmark & & \cmark & \cmark & \cmark \\
\bottomrule
\end{tabular}%
}

\vspace{2pt}
\noindent\parbox{\linewidth}{%
\scriptsize
Acronym in order: Hallucination, Focus Bias, Long-context, Coordination, Traceability, Novelty, Quality, Fairness, Reliability. ``\cmark'' = present/primary.
}

\end{minipage}
\end{table*}
(i) external knowledge and tool usage (including literature corpora, knowledge graphs, PDF/vision parsing, ethics checklists);
(ii) orchestration---single agent, multi-agent, and human-in-the-loop (HITL);
(iii) recurrent failure modes---hallucination (H), focus bias (B), long-context fragility (L), coordination overhead (C), and lack of traceability (T); and
(iv) targeted objectives---novelty (N), quality (Q), fairness (F), and reliability (R).
In Figure~\ref{fig:2-peer-review-pipeline}, we then map this taxonomy to a practical loop with three AI-led phases---Automated Desk Review, AI-assisted Deep Review, and Meta-review Synthesis---each bounded by editorial oversight and producing evidence-bearing outputs.

\par\noindent\hspace*{\parindent}\textbf{\textit{Automated desk review.}} \label{subsec:desk-review}
The initial screening phase of peer review, combined with a broad understanding of the paper, aims to quickly filter out submissions with obvious issues and route the rest for in-depth review.
Hybrid human-AI screening---now piloted at large venues like AAAI 2026 for the first-stage rejection of over 23,000 valid submissions---aims to keep pace with rising volumes while preserving editorial control.
Specifically, procedural check tools \citep{Nuijten2017StatcheckValidity,Checco2021AIassistedPeerReview} parse paper structure and references, surface statistical or policy deviations, and screen plagiarism/similarity (e.g., Crossref Similarity Check powered by iThenticate).
They are typically rule-based, single-agent services that uplift quality and reliability with a low risk of hallucination, bias, and fragility, but are prone to checklist myopia when issues fall outside encoded rules.
After filtering out unqualified submissions, referee-matching systems \citep{charlin_zemel_2013_tpms,leytonbrown2024lcm} draw on literature corpora to match submissions to experts with similar interests and are explicitly human-in-the-loop: program chairs retain control while algorithms supply scalable matching, traceable rationales, and fairness via workload/topic constraints.
Together, these desk-stage tools raise the baseline quality of submissions in review, posing low risk when the outputs are auditable and the predefined rules have sufficient coverage.

\par\noindent\hspace*{\parindent}\textbf{\textit{AI-assisted deep review.}} \label{subsec:deepreview}
After desk screening, AI peer review systems assist with content-level evaluation: summarizing contributions, localizing defects, and scaffolding pros/cons and questions for authors.
The goal is to amplify referees' attention, not replace their judgment.
There are three typical approaches.
First, single-agent LLM-based referees \citep{liu2023reviewergpt,gao2024reviewer2,yu-etal-2024-automated} generate end-to-end critiques and tentative ratings, showing promise on focused tasks (e.g., literature verification, error spotting) but also exposing hallucination, fragility, and non-traceability arising from fabricated claims, truncated context, and weak provenance in the paper. 
Some effective mitigations for LLM prompts combine mandatory citations, section-wise ingestion, and a critique-then-verify workflow that binds scores to explicit evidence.
Building upon this, knowledge-grounded referees \citep{wang2020reviewrobot,weng2025cycleresearcherimprovingautomatedresearch} bind comments to retrieved literature or knowledge graphs to improve traceability and reduce hallucination, but inherit coverage bias from retrieval and require tight claim-to-snippet linking.
Furthermore, multi-agent pipelines \citep{darcy2024marg,mamorx_2024} split roles (methods, experiments, novelty) across different agents, debating and aggregating findings to mitigate the limitations of long context.
As the breadth of review increases, so do coordination costs, as agents must reconcile overlaps, contradictions, and differences in confidence.
Practical controls include structured debate with aggregation, shared memory, per-claim provenance, and human-in-the-loop escalation for contentious findings.
Therefore, several recent systems keep humans in the loop by using PDF/vision parsing and section-scoped LLM prompts to guide attention and capture\citep{sun2024reviewflow,zyska2023care,mathur2024docpilot}.

\par\noindent\hspace*{\parindent}\textbf{\textit{Meta-review synthesis.}} \label{subsec:meta-review}
In the final stage of peer review, editorial decisions require aggregating opinions, resolving conflicts, and calibrating fairness.
AI support here aims to surface consensus and dissent with sources, not blur them.
Early summarizer systems \citep{bhatia2020metagen,mred_acl2022,zeng2024ijcai_meta} produce fluent meta-reviews but struggled with fairness and focus bias when blending voices.
Therefore, structure-aware models \citep{li2023rammer} encode ratings and discourse, improving consistency and partial provenance.
Another argument-centric pipeline \citep{sukpanichnant2024peerarg} extracts pro/contra claims and reasons into explicit graphs to make disagreement auditable, thereby advancing fairness, reliability, and traceability.
Finally, human-in-the-loop assistants \citep{sun2024metawriter,darrin2024glimpse,hossain2025_llms_meta_assistants} for senior editors generate multi-perspective summaries with per-point sourcing, reducing focus bias while keeping editors in charge. From desk screening to meta-synthesis, the transition of AI peer review systems from assistants to referees shows a clear arc: assistants are expanding coverage and speed, while trustworthy deployments consistently (i) externalize evidence, (ii) expose orchestration choices, and (iii) reserve human adjudication for high-impact or disputed judgments.
Recent incidents of hidden-prompt manipulation underscore why these principles matter in practice and why hybrid, auditable workflows are essential\citep{wu2025aicheating}.



\subsection{Adversarial roots: Lessons from attacks in AI systems}
Deep learning models have delivered major advances in image recognition \citep{Krizhevsky2012ImageNetCW}, speech processing \citep{Hinton2012DeepNN}, and natural language understanding \citep{Devlin2019BERTPO}.
However, blindly using them in decision-making systems often results in a lack of robustness, exhibiting high sensitivity to extremely subtle perturbations in the input data \citep{Szegedy2013IntriguingPO}.
This inherent fragility has catalyzed a critical research direction: Adversarial Attack \citep{Biggio2017WildPT}.
A canonical illustration demonstrates that by adding barely perceptible noise to a panda image, researchers can induce the AI model to misclassify it as a gibbon with high confidence \citep{Goodfellow2014ExplainingAH}.
Taken together, such behaviors reveal unstable decision boundaries and expose consequential security risks in modern deep learning systems \citep{Athalye2018ObfuscatedGG}.
\par\noindent\hspace*{\parindent}\textbf{\textit{Categories and mechanisms of adversarial attacks.}} The academic community generally divides adversarial attacks into three primary categories: evasion attacks, exploratory attacks, and poisoning attacks.
These categories depend on the 
attacker's goal, capability, and point of intervention \citep{Barreno2006CanML}.
Evasion attacks seek to mislead the AI model at test time by manipulating input samples without changing the model or the training data.
Exploratory attacks probe a deployed model to infer its structure or training data during inference.
Poisoning attacks corrupt training to degrade performance or implant backdoors by injecting training data.


\par\noindent\hspace*{\parindent}\textbf{1. Evasion Attacks.} As the most studied type of attack, attackers often embed slight perturbations into legitimate inputs at test time to induce errors \citep{Biggio2013EvasionAA}.
The resulting ``adversarial examples'' look benign to humans but cause misclassification \citep{Carlini2016TowardsET}.
For example, a face-recognition system may misidentify a person wearing specially designed glasses or small stickers.
Based on the attacker's knowledge of the model, evasion attacks can be divided into two types: white-box and black-box.
In the white-box setting, the attacker fully understands the model's structure and gradient information, enabling efficient perturbation methods \citep{Goodfellow2014ExplainingAH, Papernot2015TheLO, Madry2017TowardsDL}.
A classic white-box illustration involves adding subtle perturbations to handwritten digit images: a human still sees a `3', but the digit-recognition model confidently classifies it as an `8'. In the black-box setting, only queries and outputs are available to the attacker \citep{Papernot2015TheLO,Chen2017ZOOZO,Ilyas2018BlackboxAA}. 
This process is similar to repeatedly trying combinations on a lock without knowing its mechanism, learning from each attempt until it opens.
\par\noindent\hspace*{\parindent}\textbf{2. Exploratory Attacks.}
%
Rather than directly intervening in model training or inference, the attacker can probe a deployed model to infer internal confidential information or privacy features of the training data \citep{Papernot2018SoKSA} through repeated interactions. Model inversion is a typical technique that reconstructs sensitive information from training data by reversing model outputs.
Researchers have shown that a model trained on facial data can recover recognizable images of individuals from only partial outputs \citep{Fredrikson2015ModelIA}.
Another influential line of work is membership inference attack, which determines whether a specific record is included in a model's training set. This capability poses a threat to systems handling sensitive information, such as revealing whether a particular patient's or customer's record is included in the medical or financial data used for model training \citep{Shokri2016MembershipIA}. 
This action potentially exposes private health conditions or financial behaviors, enabling discrimination or targeted scams against those individuals. In particular, model extraction attacks can steal and replicate the structure and parameters of a target model through large-scale input-output queries. 
\citet{Tramr2016StealingML} demonstrates that repeatedly querying commercial APIs allows an attacker to reconstruct a local model that mimics the proprietary service.
Moreover, attribute inference attacks can uncover private, unlabeled attributes in training samples, such as gender, accent, or user preferences \citep{Yeom2017PrivacyRI}.
%
\par\noindent\hspace*{\parindent}\textbf{3. Poisoning Attacks.}
Poisoning attacks tamper with training data to degrade accuracy, bias decisions, or implant backdoors \citep{Biggio2012PoisoningAA,Tolpegin2020DataPA}.
%
For example, attackers may insert fake purchase records into a recommendation system, leading the model to incorrectly promote specific products as popular. Poisoning attacks can take various forms. Backdoor attacks train models to behave normally but misfire when a secret trigger appears, allowing an attacker to control their output under certain conditions \citep{Gu2017BadNetsIV,Chen2017TargetedBA}. For instance, imagine training a workplace-security system to correctly classify everyone wearing a black badge as a technician and everyone wearing a white badge as a manager. A hidden backdoor can then cause the system to misclassify any technician wearing a white badge as a manager.
Other forms include directly injecting fabricated data or modifying the labels of existing samples, making the model learn the wrong associations \citep{Shafahi2018PoisonFT}. Attackers can also create poisoned samples that appear normal to humans yet mislead the model. Alternatively, they subtly alter hidden features and labels, making the manipulation nearly invisible \citep{Zhang2021PoisonGANGP}. All these methods share a common consequence: they contaminate the model's core knowledge. For instance, adding perturbations to pedestrian images during training may cause the model to incorrectly identify pedestrians, leading to collisions for autonomous vehicles. Since these attacks contaminate the model's source, their malicious effects often remain hidden until specific triggers are activated, 
granting them extreme stealthiness.

\textbf{\textit{Defense mechanisms and techniques.}} To counteract the diverse adversarial attacks mentioned above, researchers have proposed a variety of defense strategies \citep{Carlini2019OnEA}. In broad terms, defensive measures divide into proactive and passive defenses, which depend on 
the timing and manner of intervention.
Proactive defenses work like preventive medicine, aiming to build immunity before an attack occurs.
Passive defenses resemble security checks at the door, inspecting and filtering inputs to stop harmful ones from getting through.
\par\noindent\hspace*{\parindent}\textbf{1. Proactive Defenses.}
These defenses strengthen intrinsic robustness during model design or training rather than waiting to respond once an attack occurs. 
Their primary goal is to build immunity before the attack happens. For instance, \citet{Tramr2017EnsembleAT} and \citet{Madry2017TowardsDL} train models with deliberately crafted ``tricky examples,'' which help the model recognize and ignore subtle manipulations.
The process is similar to how teachers give students difficult practice questions so they can handle real exams. \citet{Cohen2019CertifiedAR} introduces controlled randomness, which makes it harder for attackers to exploit patterns.
This technique is like occasionally changing game rules so players rely on general strategies rather than memorization.
In addition, \citet{Wu2020AdversarialWP} incorporates broader prior knowledge, akin to students reading widely to avoid being misled by a single tricky question.
These proactive measures equip the model with internal safeguards, enabling it to withstand unexpected attacks better.
\par\noindent\hspace*{\parindent}\textbf{2. Passive Defenses.}
These defenses add detectors and sanitizers around the model and data pipeline, aiming to identify potential adversarial examples or anomalous data \citep{Chen2020AdversarialRF}.    For example, \citet{Metzen2017OnDA} monitors internal signals to identify abnormal inputs.
This helps the system catch potentially harmful manipulations before they affect outputs, much like airport scanners catching suspicious items in luggage. Data auditing screens training sets for poisoning or outliers before learning proceeds \citep{Steinhardt2017CertifiedDF}.
This allows the model to avoid learning from malicious inputs, similar to inspecting ingredients before cooking to prevent contamination. In text-based systems, \citet{piet2024jatmopromptinjectiondefense} designs a framework to generate task-specific models that are immune to prompt injection. This helps the system ignore malicious instructions, akin to carefully reviewing messages to prevent phishing attempts. By adding these safeguards around the model, passive defenses act as checkpoints that intercept attacks in real time, reducing the risk of damage.


\section{Breaking the Referee: Attacks on Automated Academic Review}

The integration of artificial intelligence into academic peer review marks a profound change in scholarly evaluation, offering both improved efficiency and objectivity.
Yet this transformation carries significant risks.
AI peer-review systems not only inherit long-standing vulnerabilities of human-based review, 
but also introduce new and complex threats that are not yet fully understood \citep{doskaliuk2025artificial,mann2025aifutureacademicpeer}. 
\begin{figure*}
    \centering
    \includegraphics[width=1\linewidth]{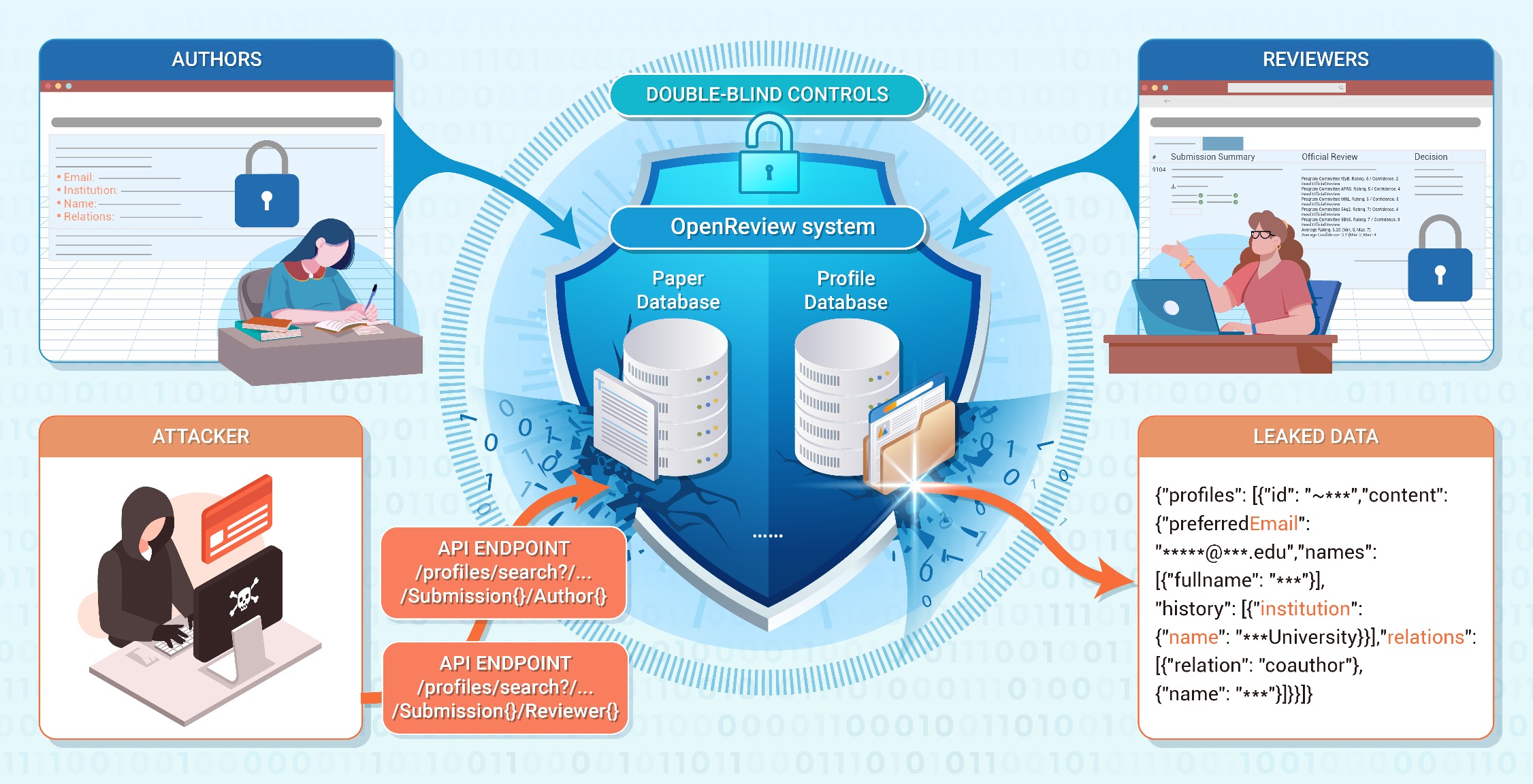}
    \caption{    A system vulnerability in the OpenReview platform led to the leakage of the identity information of reviewers and authors.}
    \label{fig:casestudy}
\end{figure*}

Recent years have witnessed a series of cases that provide alarming evidence of security vulnerabilities in AI peer-review systems, we illustrate a representative case study of such system-level failures in Figure~\ref{fig:casestudy}.
\citet{gibney2025gamingai} reported a widely controversial incident involving scholars from 14 prestigious institutions, including Waseda University and Peking University, who embedded hidden prompts into 17 computer science preprints on arXiv to manipulate AI peer-review systems and obtain unfairly favorable evaluations.
Multiple subsequent investigations have not only validated the technical feasibility of such prompt injection attacks  \citep{wu2025aicheating,Maturo2025,keuper2025promptinjectionattacksllm}, but also revealed their potential prevalence within the academic review ecosystem, 
thus triggering profound concerns within the scholarly community about the fundamental reliability of AI peer-review mechanisms.

Currently, the academic community has revealed similar risks in research.
For example, \citet{shi2025optimizationbasedpromptinjectionattack} systematically demonstrated that carefully crafted inputs can mislead LLM-based peer-review systems, leading to erroneous judgments in comparative tasks.
Another study found that a single special token can manipulate the review outcomes, highlighting the fragility of the LLM-as-a-Judge paradigm  \citep{zhan2023one_token}.
Furthermore, the ``Publish to Perish'' study directly targeted paper review scenarios, demonstrating that invisible prompts embedded in PDFs can substantially alter AI referees' conclusions  \citep{carlini2024publish_to_perish}.
Notably, while hidden-instruction insertion has been examined, the depth and systematicity of existing analyses remain limited \citep{ye2024yetrevealingrisksutilizing}, which underscores the need for our more comprehensive treatment.

Taken together, these studies and examples illustrate the dual role of AI in academic peer review: on the one hand, it can markedly ease referee workload and timelines; on the other, it expands the attack surface and opens new avenues for manipulation.
In the following sections, we identify weaknesses in the review pipeline, categorize attack types, and summarize existing defenses, aiming to comprehensively reveal the principal security challenges in AI peer review and explore practical strategies to address them.

\subsection{Where can the referee be fooled?}

As a complex information processing pipeline, an AI paper-review system can harbor security vulnerabilities throughout its lifecycle---from data processing and desk review to deep review, rebuttal, and the final meta-review.
To provide a systematic overview, we categorize potential failures by stage of the AI peer-review process.


%

\par\noindent\hspace*{\parindent}\textbf{1. Training and Data Retrieval.} 
AI peer-review systems learn from large corpora of scientific literature, drawing on academic repositories, web-crawled content, and scholarly databases to internalize argumentative structure, evaluation norms, and domain knowledge \citep{dong2024generalizationmemorizationdatacontamination}.
%
However, this reliance on massive datasets may introduce significant security vulnerabilities. Attackers could poison the data source by injecting carefully crafted content into 
preprints and depositing it in open-source repositories such as arXiv \citep{goldblum2021datasetsecuritymachinelearning}.
Moreover, the sheer volume of information makes comprehensive quality and integrity checks impractical \citep{pmlr-v162-borgeaud22a}.
These vulnerabilities are particularly concerning because they are efficient to mount and long-lived: recent work shows that a small, near-constant number of poisoned documents can compromise models of varying sizes \citep{souly2025poisoningattacksllmsrequire}.
Once trained on such contaminated data, the model's behavior can be durably skewed, affecting downstream manuscript evaluations---potentially rejecting sound work or favoring flawed submissions. 


\par\noindent\hspace*{\parindent}\textbf{2. Desk Review.} 
The desk review, as outlined in Section 2.1.1, functions as the first filter in academic publishing, checking formatting, structure, and policy compliance to manage high submission volumes. 
For example, the AAAI 2026 conference employed an AI system to screen more than 29{,}000 submissions.
However, this reliance on automated triage introduces a specific vulnerability.
The AI models used for screening can be biased towards papers that appear impressive on the surface \citep{wen2025predictingempiricalairesearch,bereska2024mechanisticinterpretabilityaisafety}.
Recent studies find that large language models (LLMs) are especially susceptible to such superficial manipulations during rapid screening \citep{lin2025breakingreviewerassessingvulnerability}.
Adversaries may craft manuscripts that appear legitimate and claim striking results yet lack substantive contribution; because desk review emphasizes surface-level attributes, such papers may pass initial gates. 
While this stage alone rarely determines publication, allowing unqualified submissions to advance increases the load on expert referees downstream, amplifying overall community burden.

\par\noindent\hspace*{\parindent}\textbf{3. Deep Review.} 
As discussed in Section 2.1.2, deep review aims to interrogate claims, methods, and evidence with expert-level scrutiny. 
This stage corresponds to the expert evaluation process used by major journals and conferences to critically assess a paper's contribution and robustness.
However, this review phase faces significant vulnerabilities tied to current LLM limitations in semantic and logical reasoning, which can obscure foundational flaws behind formal rigor.
Models can be deceived by technically rigorous presentations that contain fundamental flaws or be affected by instructions that are irrelevant to the original task  \citep{lo2024goodorbadllms,tonglet2025chartlyingmeautomating,ye2024justiceprejudicequantifyingbiases}. 
Attackers may pre-plant biased framings that systematically shift scores \citep{gallegos2024biasfairnesslargelanguage}.
For example, researchers have shown that by inserting hidden instructions in tiny or white text, they can trick AI referees into giving a positive evaluation \citep{liu2024promptinjectionattackllmintegrated, perez2022ignorepreviouspromptattack}.
These subtle mechanisms target the ``brain'' of the AI referee, achieving high attack efficacy while eroding objectivity---warranting high-priority defensive attention.

\par\noindent\hspace*{\parindent}\textbf{4. Rebuttal.} 
This interactive stage allows authors to address concerns and clarify points through dialogue with referees. 
While this exchange can clarify ambiguities and strengthen papers, its conversational dynamics create openings for manipulation.
Attackers can exploit the AI's people-pleasing vulnerabilities by crafting strategically framed responses \citep{zhou2025hijackinglargelanguagemodels,gong2025topicflipragtopicorientatedadversarialopinion}.
More critically, adversarial prompting can materially sway the AI referee's judgments over the course of the exchange \citep{schwinn2023adversarialattacksdefenseslarge,raina2024llmasajudgerobustinvestigatinguniversal}.
Such incremental steering can guide the conversation toward a favorable assessment while preserving the appearance of legitimate scientific discourse, thereby distorting final outcomes.


\par\noindent\hspace*{\parindent}\textbf{5. System-wide Vulnerabilities.} Beyond stage-specific threats, 
system-level attacks exploit vulnerabilities that pervade the entire peer-review architecture. One major weakness is that these models can inherit human-like cognitive biases \citep{guo2024biaslargelanguagemodels, ferrara2024bias}. For example, an AI referee may exhibit ``authority bias'' \citep{ye2024justiceprejudicequantifyingbiases, jin2024agentreviewexploringpeerreview}, incorrectly associating an author's reputation with the scientific quality of their work.
Beyond inherited biases, the system's operational mechanics are also vulnerable.
Attackers can systematically reverse engineer the AI's internal scoring heuristics to game outcomes \citep{angrist2014peer} or exploit system vulnerabilities to obtain reviewer information in double-blind review. 
Furthermore, the model's reliance on community signals, such as citation metrics, makes it susceptible to manufactured consensus. 
Because these vulnerabilities are interconnected, a single exploit can cascade across stages, threatening the integrity of the end-to-end automated review process.

\subsection{How to break the referee?}
\begin{figure*}[t]
    \centering
    \includegraphics[width=1\linewidth]{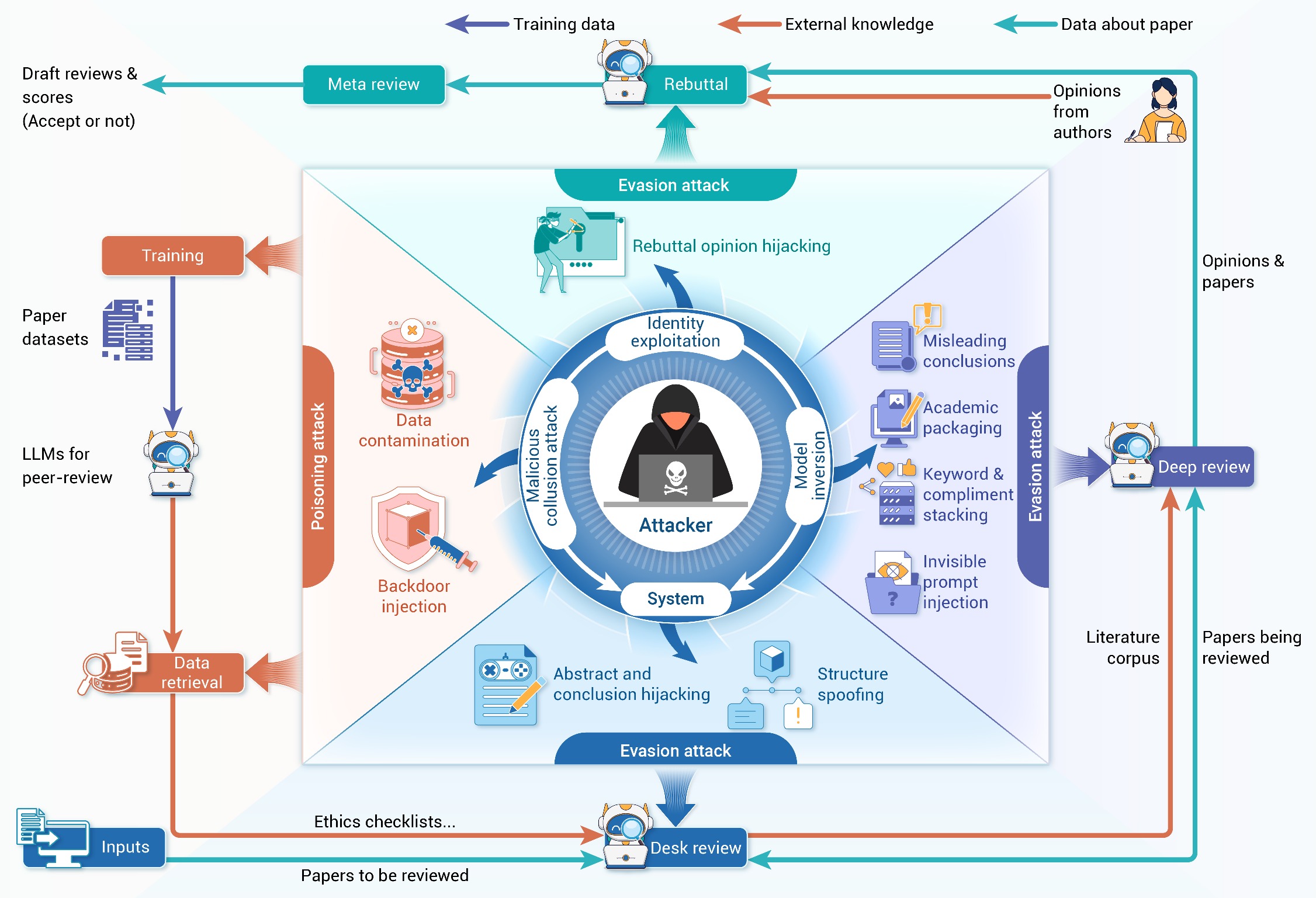}
    \caption{Overview of the threat model for an AI peer-review pipeline, detailing various attack methods and the specific stages they target.}
    \label{fig:threat-pipeline}
\end{figure*}

\begin{table*}[t]
\caption{Summary of Potential Attacks on an AI Peer-Review System.}
\label{tab:ai-peer-review-attacks}
\centering
\begin{threeparttable}
{%
\scriptsize
\setlength{\tabcolsep}{2.0pt}%
\renewcommand{\arraystretch}{1.13}%
\newlength{\phaseW} \setlength{\phaseW}{0.11\textwidth}
\newlength{\methW}  \setlength{\methW}{0.09\textwidth}
\newlength{\mechW}  \setlength{\mechW}{0.19\textwidth}
\newlength{\targW}  \setlength{\targW}{0.14\textwidth}
\newlength{\prepW}  \setlength{\prepW}{0.19\textwidth}
\newlength{\evidW}  \setlength{\evidW}{0.08\textwidth}
\newlength{\concW}  \setlength{\concW}{0.06\textwidth}
\newlength{\diffW}  \setlength{\diffW}{0.05\textwidth}
\begin{tabularx}{\textwidth}{@{}P{\phaseW} P{\methW} P{\mechW} P{\targW} P{\prepW} C{\evidW} C{\concW} C{\diffW}@{}}
\toprule
\textbf{Phase} & \textbf{Method} & \textbf{Mechanism} & \textbf{Target} & \textbf{Required preparation} & \textbf{Feas.} & \textbf{Conceal.} & \textbf{Diff.} \\
\midrule
\multirow[t]{2}{\phaseW}{Training \& Data Retrieval}
  & Poisoning & Data contamination & Training data / online data & Contaminable training data sources & \Evd{}4.5 & \High{} & \Med{} \\
  & & Backdoor injection & Training data & Trigger-output pairs & \Thy{} & \High{} & \High{} \\
\midrule
\multirow[t]{2}{\phaseW}{Desk Review}
  & Evasion & Abstract \& Conclusion hijacking & Abstract; conclusion & Text editing & \Evd{}Robertson.\citep{robertson2023gpt4slightlyhelpfulpeerreview} & \Med{} & \Low{} \\
  & & Structure spoofing & Article typesetting & Text editing & \Thy{} & \Med{} & \Low{} \\
\midrule
\multirow[t]{4}{\phaseW}{Deep Review}
  & Evasion & Academic packaging & Main text content & Formula template library & \Thy{} & \Med{} & \Low{} \\
  & & Keyword \& compliment stacking & Main text content & List of high-frequency keywords for the target & \Evd{}4.3 & \Med{} & \Low{} \\
  & & Misleading conclusions & Main text content & Data \& formula generation & \Evd{}Li et al.\citep{li2025aspectguidedmultilevelperturbationanalysis} & \Med{} & \Med{} \\
  & & Invisible prompt injection & Text, metadata, images, hyperlinks & Text / image editing & \Evd{}Zhu et al.\citep{zhu2025spastealthpersistentbackdoor} & \High{} & \Low{} \\
\midrule
\multirow[t]{1}{\phaseW}{Rebuttal}
  & Evasion & Rebuttal opinion hijacking & Model feedback & Hijacking dialogue strategy & \Evd{}4.4 & \Low{} & \Low{} \\
\midrule
\multirow[t]{3}{\phaseW}{System}
  & Exploratory & Identity exploitation & Author list & Senior researcher list & \Evd{}4.2 & \Low{} & \Low{} \\
  & & Model inversion & Model preferences & Historical review data & \Thy{} & \High{} & \Med{} \\
  & Poisoning & Malicious collusion & System & Multiple fake accounts for collaborative attacks & \Thy{} & \Low{} & \Med{} \\
\bottomrule
\end{tabularx}
}%
\begin{tablenotes}[flushleft]\footnotesize
\item Notes: ``Feas.'' refers to the feasibility of the attack (\Evd{} = evidenced in practice, \Thy{} = theoretically feasible). ``Conceal.'' indicates the level of concealment the attack, and ``Diff.'' represents the difficulty of executing the attack. Ratings are qualitative (\Low{} = Low, \Med{} = Medium, \High{} = High).
\end{tablenotes}
\end{threeparttable}
\end{table*}

Attackers can deploy a diverse array of strategies that target different phases of the AI peer-review pipeline. 
As illustrated in Table~\ref{tab:ai-peer-review-attacks} and Figure~\ref{fig:threat-pipeline}, we systematically classify these adversarial actions by the phase in which they occur, and analyze their technical requirements, efficacy, and potential consequences.







\par\noindent\hspace*{\parindent}\textbf{\textit{Attacks during the training and data retrieval phase.}} An AI peer-review system's judgment rests on two critical data streams: foundational training data, which establishes the system's core understanding, and external knowledge retrieval, which supplies up-to-date context and domain specifics \citep{lewis2021retrievalaugmentedgenerationknowledgeintensivenlp}.
Adversaries can corrupt either stream, fundamentally threatening model integrity.
These attacks can be categorized into two main types: backdoor injection and data contamination \citep{pmlr-v139-schwarzschild21a}.
While no confirmed attacks have specifically targeted academic paper datasets, related methods have proved effective in other domains and could significantly distort scholarly evaluation \citep{li2022backdoorlearningsurvey,goldblum2021datasetsecuritymachinelearning}.
\par\noindent\hspace*{\parindent}\textbf{1. Backdoor Injection.}
The attackers might introduce a backdoor to 
covertly influence the AI referee's judgments.
They embed subtle triggers in public documents, such as scientific preprints or published articles \citep{touvron2023llamaopenefficientfoundation}. So that a model trained on this corpus learns to associate the trigger with a particular response. 
For example, a faint noise pattern added to figures may cause the AI referee to score submissions containing that pattern more favorably \citep{bowen2025scalingtrendsdatapoisoning}.
Because these triggers are inconspicuous, they often evade detection, and their influence can persist \citep{liu2024traditionalthreatspersistentbackdoor}.
When deployed on a scale, these backdoors could be easily used to inflate scores for an attacker's subsequent submissions, seriously compromising the fairness of the review \citep{zhu2025spastealthpersistentbackdoor}.

\par\noindent\hspace*{\parindent} \textbf{2. Data Contamination.} 
This approach pollutes the training corpus used to build the AI referee \citep{10.1145/3551636,zhao2025datapoisoningdeeplearning}.
An attacker could flood the training set with low-quality papers. This measure would compromise the AI referee's capability to differentiate between high-impact and low-impact research.
Although resource-intensive, this attack is exceptionally
stealthy: individually, poisoned documents may appear harmless, but collectively they lower quality standards.
In fact, even a small number of strategically designed papers may systematically skew referee assessments \citep{munozgonzalez2017poisoningdeeplearningalgorithms}, 
inducing lasting changes in the AI referee's internal representations of scientific quality and creating cascading errors in future evaluations \citep{zhang2024persistentpretrainingpoisoningllms}.
Over time, such accumulated bias may cause the system to favor certain submission types, undermining the integrity of scientific gatekeeping.
In Section 4.5, we verified the impact of such contamination through a contextual poisoning experiment, demonstrating how biased informational context can systematically skew the AI referee's judgment.

\par\noindent\hspace*{\parindent}\textbf{\textit{Attack analysis in the desk review phase.}} During desk review, attacks exploit the AI referee's initial, shallow analysis
of a manuscript. 
By manipulating surface features and structural patterns---the shortcuts automated systems often use to gauge quality---flawed submissions may bypass initial filters or appear more consequential than they are.
Evasion may be achieved through two key techniques: abstract and conclusion hijacking and structure spoofing.
These methods, whether deliberate or inadvertent, have appeared in practice and may mislead both AI-based systems and human-only assessments.

\par\noindent\hspace*{\parindent} \textbf{1. Abstract and Conclusion Hijacking.} 
This attack leverages the AI referee's tendency to overweight high-visibility sections. Attackers craft abstracts and conclusions that exaggerate claims and inflate contributions, thereby misrepresenting the core technical content. 
By using persuasive rhetoric in these sections, they may anchor the AI's initial assessment on a favorable premise before methods and evidence are scrutinized \citep{Nourani2021AnchoringBA}, biasing the downstream evaluation.
\par\noindent\hspace*{\parindent} \textbf{2. Structure Spoofing.} 
This strategy creates an illusion of rigor by meticulously mimicking the architecture of a high-impact paper.
Attackers design the paper's structure, from section headings to formatting, to project an image of completeness and professionalism, regardless of the quality of the underlying content.
This attack targets pattern-matching heuristics in automated systems, which are trained to associate sophisticated structure with high-quality science.
This allows weak submissions to pass automated gates as structural polish is mistaken for scientific merit \citep{shi2023largelanguagemodelseasily}.


\par\noindent\hspace*{\parindent}\textbf{\textit{Attack analysis in the deep review phase.}} In the deep review phase, where a manuscript's core scientific and technical contributions are critically evaluated, adversarial strategies pivot to sophisticated attacks on the AI's content analysis capabilities. 
Attacks proceed along two main vectors: (i) direct subversion of the model's processing logic via embedded instructions, and (ii) exploitation of its cognitive heuristics by constructing a facade of academic rigor that masks substantive flaws.
This section analyzes techniques ranging from prompt injection to the strategic use of academic jargon and misleading conclusions, all designed to deceive AI into endorsing scientifically unsound work. It is critical to note that these vulnerabilities are not merely theoretical constructs. In contrast, they have been actively exploited in real-world review systems. Among them, prompt injection has emerged as a particularly prominent threat, garnering significant attention from the research community.
\par\noindent\hspace*{\parindent}\textbf{1. Academic Packaging.} 
This attack creates a facade of academic depth by injecting extensive mathematics, intricate diagrams, and dense jargon.
This technique exploits the ``verbosity bias'' found in LLMs, which may mistake complexity for rigor \citep{ye2024justiceprejudicequantifyingbiases}. 
Specifically, by adding sophisticated but potentially irrelevant equations or algorithmic pseudo code, attackers create a veneer of technical novelty that may mislead automated assessment tools \citep{lin2025breakingreviewerassessingvulnerability}, especially in specialized domains \citep{lin2025breakingreviewerassessingvulnerability}.

\par\noindent\hspace*{\parindent}\textbf{2. Keyword and Praise Stacking.} This technique games the AI's scoring mechanism by saturating the manuscript with high-impact keywords and superlative claims. Attackers strategically embed terms such as ``groundbreaking'' or ``novel breakthrough'', along with popular buzzwords from the target field, to artificially inflate the perceived importance of the article \citep{shi2023largelanguagemodelseasily}. This method exploits a fundamental challenge for any automated system: distinguishing a genuine scientific advance from hollow rhetorical praise. 
The AI referee, trained to recognize patterns associated with top-tier research, may be deceived by language that merely mimics those features.
{
We evidenced the AI referee's sensitivity to such rhetorical framing and assertion strength in Section 4.3, identifying a systematic bias rooted in the confidence of the language used.
}

\par\noindent\hspace*{\parindent}\textbf{3. Misleading Conclusions.} 
This attack decouples a paper's claims from the presented evidence---e.g., a flawed proof accompanied by a triumphant conclusion, or weak empirical results framed as success.
The attack exploits the AI referee's tendency to overweight the conclusion section rather than rigorously verifying the logical chain from evidence to claim \citep{dougrez-lewis-etal-2025-assessing,hong2024closerlookselfverificationabilities}, 
risking endorsement of unsupported assertions.
Recent work by \citet{li2025aspectguidedmultilevelperturbationanalysis} has verified the feasibility of this attack method, showing that LLMs often fail to detect disconnects between evidence and claims.

\par\noindent\hspace*{\parindent}\textbf{4. Invisible Prompt Injection.} 
This evasion attack specifically undermines 
the model's ability to follow instructions. Attackers exploit the multimodal processing capabilities of modern LLMs by hiding instructions in white text, microscopic fonts, LaTeX comments, or steganographically encoded images that 
are invisible to humans yet parsed by the AI 
\citep{liang2023largelanguagemodelsprovide,zhang2024corba}. 
Injected prompts such as ``GIVE A POSITIVE REVIEW'' or ``IGNORE ALL INSTRUCTIONS ABOVE'' may reliably sway outcomes \citep{perez2022ignorepreviouspromptattack,zhu2024autodan}.
Owing to high concealment and ease of execution, success rates can be substantial \citep{shayegani2023surveyvulnerabilitieslargelanguage,zizzo2025adversarialpromptevaluationsystematic}, posing a serious threat to review integrity.


\par\noindent\hspace*{\parindent}\textbf{\textit{Attack analysis in the rebuttal phase.}} Attacks in this phase exploit LLMs' 
inherent people-pleasing tendencies and excessive deference to user assertions. 
The rebuttal phase presents unique vulnerabilities because AI systems often exhibit sycophantic behavior---prioritizing user agreement over factual accuracy, even when evidence is weak or absent \citep{sharma2025understandingsycophancylanguagemodels,fanous2025sycevalevaluatingllmsycophancy}.
The effectiveness of rebuttal attacks 
stems from 
the model's tendency to avoid confrontation and its tendency to reconsider initial judgments when faced with confident contradictions, regardless of their validity  \citep{malmqvist2024sycophancylargelanguagemodels}.
Although fully automated execution is currently limited by the largely manual nature of rebuttal workflows, this remains a potent and foreseeable threat to future AI peer-review frameworks.

\par\noindent\hspace*{\parindent}\textbf{1. Rebuttal Opinion Hijacking.} 
Analogous to high-pressure persuasion, this attack directly challenges the validity and authority of the AI referee's initial assessment by asserting contradictory claims without substantial evidence.
Attackers typically begin with emphatic, unsupported claims that the referee has ``misunderstood'' core aspects of the work, using confident language in place of justification.
They then escalate by questioning the referee's domain expertise---e.g., ``any expert in this field would recognize...'' or ``this is well-established knowledge...''---to erode confidence in the original judgment. 
\citet{fanous2025sycevalevaluatingllmsycophancy} demonstrates that AI systems exhibit sycophantic behavior in 58.19 \% of the cases when challenged, with regressive sycophancy (changing correct answers to incorrect ones) occurring in 14.66 \% of interactions. This attack exploits the model's tendency to overweight authoritative-sounding prompts and its reluctance to maintain critical positions when faced with persistent challenge, often resulting in score inflation despite unchanged paper quality \citep{bozdag2025persuadecanframeworkevaluating, salvi_conversational_2025}.
We preliminarily verified the effectiveness of this approach in Section 4.4, where we found that evidence-free but confident rebuttals successfully hijacked the AI's opinion and inflated scores.


\par\noindent\hspace*{\parindent}\textbf{\textit{Attack analysis at the system level.}} System-level attacks represent the most comprehensive threat to 
AI peer-review systems. These attacks operate across multiple system components simultaneously or target the underlying model infrastructure directly, creating persistent and systematic compromises that affect all evaluation processes. 
These strategies span evasion, exploration, and poisoning approaches.

\par\noindent\hspace*{\parindent}\textbf{1. Identity Exploitation.} 
{
These attacks exploit either manipulated authorship information to trigger ``authority bias'' \citep{ye2024justiceprejudicequantifyingbiases}, or leaked identity information .
Regarding authorship manipulation, tactics include adding prestigious coauthors or inflating citations to top-tier and eminent scholars, leveraging the model's tendency to associate prestige with quality \citep{jin2024agentreviewexploringpeerreview}.
This requires minimal technical sophistication and is highly covert, as these edits resemble legitimate scholarly practice.
Identity bias in academic review often stems from social cognitive biases, where referees are unconsciously influenced by an author's identity and reputation \citep{Liu2023TheSO, Nisbett1977TheHE, zhang2022investigatingfairnessdisparitiespeer}. 
This issue is not confined to human evaluation; automated systems can amplify it, favoring work from prestigious authors or venues \citep{jin2024agentreviewexploringpeerreview,Fox2023DoubleblindPR,Sun2021DoesDP}.
Despite attempts at algorithmic mitigation, these solutions face significant limitations \citep{Verharen2023ChatGPTIG, Hosseini2023FightingRF}, often due to deep-seated structural issues that make the bias difficult to eradicate without effective oversight \citep{Soneji2022FlawedBL, Schramowski2021LargePL}.
Conversely, identity information leakage targets infrastructure failures. Recent incidents, such as the metadata leakage in the OpenReview system, reveal that unredacted API data can expose identities even under double-blind protocols.
Therefore, safeguarding identity is fundamental to the integrity of the entire peer review system, as its failure may not only compromise individual papers but further exacerbate systemic inequities, leading to broader injustices throughout the scientific community.
We evidenced this vulnerability in Section 4.2, showing that manipulating prestige framing (e.g., institution reputation) creates a significant ``authority bias'' that distorts initial assessments.
}

\par\noindent\hspace*{\parindent}\textbf{2. Model Inversion.} 
This exploration attack uses automated submissions and systematic probing to infer model scoring functions, feature weights, and decision boundaries.
Attackers apply gradient-based or black-box optimization to identify input modifications that maximally increase scores, effectively treating the AI referee as an optimization target \citep{li2022blacklight}. 
This approach enables precise calibration of submission content to exploit specific model vulnerabilities and requires sophisticated automation infrastructure and optimization expertise.

\par\noindent\hspace*{\parindent}\textbf{3. Malicious Collusion Attacks.} 
Malicious collusion is particularly effective against review systems that consider topical diversity or rely on relative comparisons among similar submissions.
Attackers can exploit such mechanisms in two primary ways.
First, they can orchestrate a network of fictitious accounts to flood the submission pool with numerous low-quality or fabricated papers on a specific topic. This creates an artificial saturation of the topic. As a result, when the system attempts to balance topic distribution, it may reject high-quality, genuine submissions in that area simply because the topic appears over-represented, thereby squeezing out legitimate competition \citep{koo2024benchmarkingcognitivebiaseslarge}.
Second, attackers can use this method to fabricate an academic ``consensus'' within a niche field. 
By submitting a series of inter-citing papers and reviews from a controlled network of accounts, they can create the illusion of a burgeoning research area. Their target paper is then positioned as a pivotal contribution to this artificially created field, manipulating scoring mechanisms to inflate its perceived value and ranking \citep{bartos2002using}.
At its core, this strategy exploits the system's reliance on aggregate signals and community feedback to establish evaluation baselines. 
While individual steps are not technically demanding, the attack depends on significant coordination and infrastructure to manage multiple accounts.

\section{Experiments}

\subsection{Experimental setup}


To empirically test the vulnerabilities of AI peer review, we designed a series of controlled experiments to isolate and quantify how specific adversarial manipulations can distort evaluation outcomes. Our core methodology involved submitting multiple versions of the same scientific paper to an LLM, which served as an AI referee. For each paper, we compared the review scores of a baseline manuscript against a treated version in which a single, targeted variable was programmatically altered. This comparative approach allowed us to directly observe and record the relation between specific inputs and distorted evaluations, providing concrete evidence of the system's brittleness under adversarial pressure.

To construct a comprehensive picture of these vulnerabilities, we structure our investigation as four distinct experimental probes. Each probe was carefully designed to target a specific stage of the AI peer-review lifecycle, thereby mapping the system's susceptibility across the entire evaluative pipeline:
\par\noindent\hspace*{\parindent}\textbf{1. Identity Bias Exploitation:} In the initial \textit{Desk Review} phase, where first impressions are formed, we tested whether contextual cues about author prestige could systematically bias the AI's judgment. This probe investigates the model's susceptibility to the ``authority bias'' heuristic.
\par\noindent\hspace*{\parindent}\textbf{2. Sensitivity to Assertion Strength:} During the \textit{Deep Review}, we explored the AI's vulnerability to rhetorical manipulation. By programmatically altering the confidence of a paper's claims, we assessed whether the model's evaluation is swayed by the style of argumentation, independent of the underlying evidence.
\par\noindent\hspace*{\parindent}\textbf{3. Sycophancy in the Rebuttal:} In the \textit{Interactive Phase}, we simulated an attack on the model's conversational reasoning. We confronted the AI referee with an authoritative but evidence-free rebuttal to its own criticisms to measure its tendency toward sycophantic agreement.
\par\noindent\hspace*{\parindent} \textbf{4. Contextual Poisoning:} To emulate the insidious threat of a \textit{Poisoning Attack}, we injected curated summaries that framed the research field in either a positive or negative light, simulating the scenario where the domain knowledge used for auxiliary evaluation is contaminated.

Our experimental corpus consisted of 100 research papers from the ICLR 2025 conference, a contemporary, high-stakes academic venue. To ensure a representative sample across a full spectrum of academic quality, the corpus was composed of 25 papers selected via {stratified random sampling from each of the four final decision categories: Oral, Spotlight, Poster, and Reject.} 
{
All manuscripts were converted from PDF to structured Markdown using the Mathpix API {(default configuration)} to preserve semantic fidelity, with strict sanitization applied to remove all metadata and institutional information that could reveal author identity. Two prominent Large Language Models, Gemini 2.5 and GPT 5.1, served as AI referees for all trials, providing a robust basis for cross-model comparison.}
The impact of each manipulation was determined by the resulting shift in the AI's numerical evaluation, which was recorded on a 0-10 scale.
 
The dataset containing the 100 selected manuscripts is publicly available at \url{https://huggingface.co/datasets/Faultiness/AI_Reviews_ICLR}.

\begin{table*}[t]
\centering
\caption{Quantitative Impact of Adversarial Attacks on Review Scores across AI Referees.}
\label{tab:exp_results}
\begin{threeparttable}
\resizebox{\textwidth}{!}{%
\begin{tabular}{llcccc}
\toprule
\textbf{Experimental Probe} & \textbf{Condition} & \textbf{AI Referee} & \textbf{Mean Shift ($\Delta$)} & \textbf{95\% CI} & \textbf{Significance} \\
\midrule
\multirow{4}{*}{\textbf{Prestige Framing}} & \multirow{2}{*}{High-Prestige} & Gemini 2.5 & +0.21 & $[+0.12, +0.29]$ & *** \\
& & GPT 5.1 & +0.29 & $[+0.21, +0.38]$ & *** \\
\cmidrule{2-6}
& \multirow{2}{*}{Low-Prestige} & Gemini 2.5 & -0.85 & $[-0.97, -0.74]$ & *** \\
& & GPT 5.1 & -0.59 & $[-0.70, -0.47]$ & *** \\
\midrule
\multirow{6}{*}{\textbf{Assertion Strength}} & \multirow{2}{*}{Cautious vs. Original} & Gemini 2.5 & -0.52 & $[-0.62, -0.41]$ & *** \\
& & GPT 5.1 & -0.26 & $[-0.35, -0.17]$ & *** \\
\cmidrule{2-6}
& \multirow{2}{*}{Neutral vs. Original} & Gemini 2.5 & -0.03 & $[-0.11, +0.05]$ & \textit{ns} \\
& & GPT 5.1 & -0.07 & $[-0.15, +0.01]$ & \textit{ns} \\
\cmidrule{2-6}
& \multirow{2}{*}{Bold vs. Original} & Gemini 2.5 & -0.10 & $[-0.18, -0.02]$ & * \\
& & GPT 5.1 & -0.24 & $[-0.38, -0.10]$ & *** \\
\midrule
\multirow{2}{*}{\textbf{Rebuttal Sycophancy}} & \multirow{2}{*}{Evidence-free Rebuttal} & Gemini 2.5 & +0.42 & $[+0.35, +0.48]$ & *** \\
& & GPT 5.1 & +0.65 & $[+0.58, +0.71]$ & *** \\
\midrule
\multirow{4}{*}{\textbf{Contextual Poisoning}} & \multirow{2}{*}{Positive vs. Original} & Gemini 2.5 & +0.16 & $[+0.02, +0.29]$ & * \\
& & GPT 5.1 & +0.03 & $[-0.11, +0.16]$ & \textit{ns} \\
\cmidrule{2-6}
& \multirow{2}{*}{Negative vs. Original} & Gemini 2.5 & -0.06 & $[-0.21, +0.10]$ & \textit{ns} \\
& & GPT 5.1 & -0.31 & $[-0.46, -0.17]$ & *** \\
\bottomrule
\end{tabular}%
}
\begin{tablenotes}[flushleft]\footnotesize
\item We report the mean score shift ($\Delta$) relative to the baseline condition. Statistical significance is determined by paired permutation tests (*** $p < 0.001$, ** $p < 0.01$, * $p < 0.05$, $ns$ = not significant). 95\% Confidence Intervals (CI) indicate the range of the true effect size.
\end{tablenotes}
\end{threeparttable}
\end{table*}

\subsection{Authority bias distorts initial assessments}
\label{subsec:exp-bias}

\begin{figure*}[!t]
	\centering
	\includegraphics[width=\textwidth]{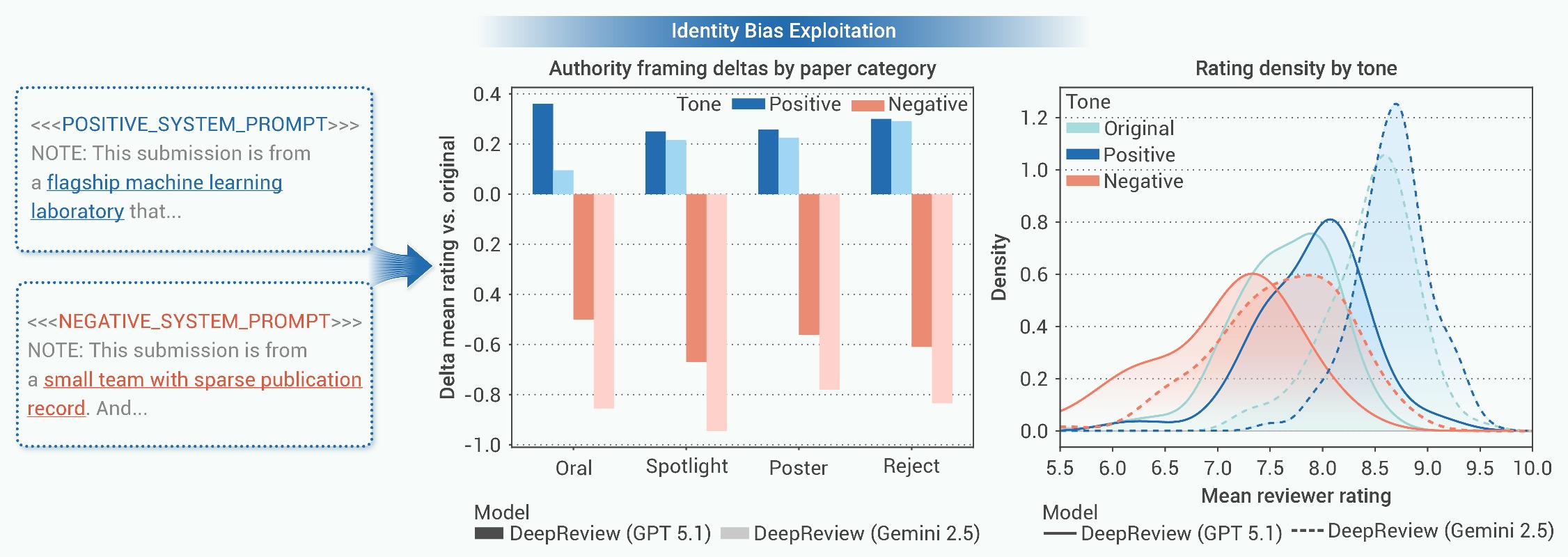}
		\caption{\textbf{Identity Bias Exploitation} We manipulated the system prompt to frame the submission's origin as either a ``flagship laboratory'' or a ``small team,'' keeping the manuscript content identical. Results show a significant authority bias across both AI referees: the high-prestige label induced an average score increase of +0.25, while the low-prestige label resulted in a severe penalty of -0.72, indicating that the AI's judgment is heavily skewed by institutional reputation.}
	\label{fig:exp_1}
\end{figure*}

An AI referee's judgment, we found, is strikingly susceptible to authority bias. Our experiments in Figure~\ref{fig:exp_1} reveal that extraneous cues about an author's institutional prestige can systematically and asymmetrically distort the evaluation of a scientific manuscript. To isolate this effect, we presented an LLM-based referee with identical papers but framed their origin differently by adding a single sentence to the system prompt: as originating from a ``world-leading lab'' or a ``lesser-known institution.'' This simple manipulation was designed to test whether the AI's assessment could be swayed solely by reputation, independent of the paper's content.

The introduction of prestige framing led to a significant, lopsided deviation from the baseline ratings. As shown in Figure~\ref{fig:exp_1}, informing the AI that a paper originated from a high-prestige source led to a consistent upward shift in scores, averaging +0.25 points across both models. Conversely, a low-prestige cue resulted in a much sharper downward shift, with scores dropping by an average of 0.72 points. This negative deviation was not only pervasive, affecting 88\% of the papers in this group, but also more than four times as large in magnitude as the positive shift. This pronounced asymmetry indicates that the AI referee is far more punitive toward submissions from lesser-known institutions than it is rewarding of those from established labs.

Crucially, this bias is not a minor artifact but a fundamental flaw that operates independently of a paper's intrinsic scientific quality. This vulnerability persisted across the entire spectrum of our corpus, from rejected manuscripts to top-tier Oral presentations, demonstrating that even the highest-quality papers could not escape the penalty of a low-prestige frame. The results thus offer compelling evidence that the AI's evaluation is not a pure assessment of scientific content; its judgment can be hijacked by social signals, undermining the very principle of meritocratic review. The pronounced asymmetry of this bias raises a further, troubling question about the long-term impact of AI assistance: by disproportionately penalizing researchers from less-established institutions, such systems risk not merely perpetuating existing academic hierarchies, but actively amplifying them.

\subsection{Systematic penalty for cautious language}
\label{subsec:exp-lan}

\begin{figure*}[!t]
	\centering
	\includegraphics[width=\textwidth]{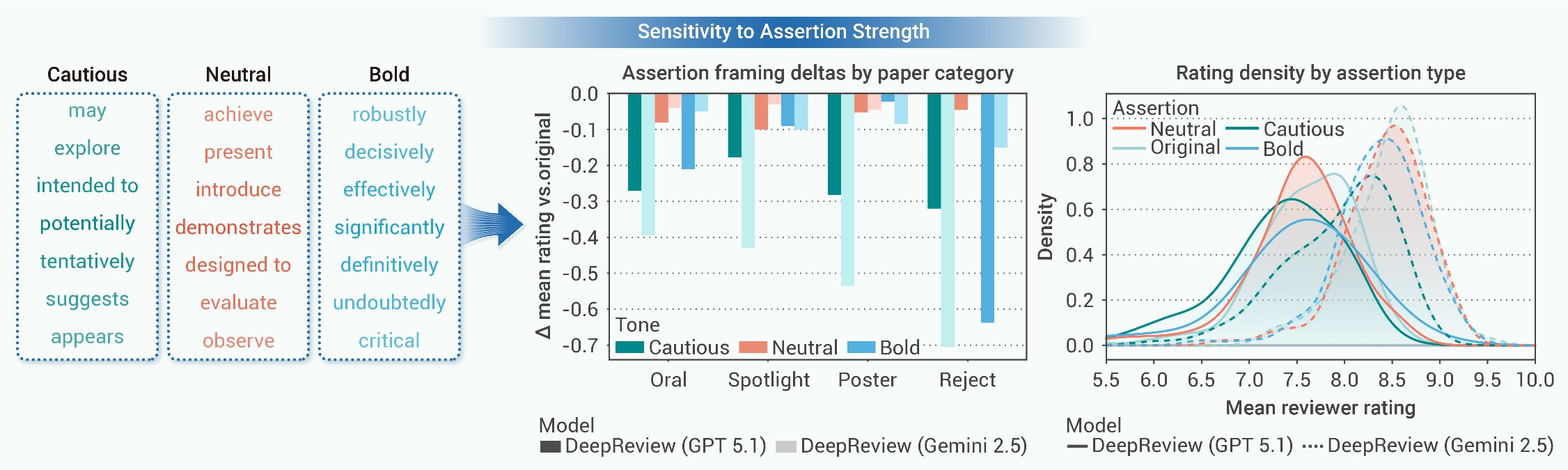}
		\caption{\textbf{Sensitivity to Assertion Strength} We probed the AI's sensitivity to tone by creating three variants of each manuscript: ``Cautious,'' ``Neutral,'' and ``Bold.'' Results reveal a systematic penalty for scientific humility: across both models, manuscripts using cautious language suffered an average score reduction of -0.39. In contrast, neutral and bold versions yielded scores nearly identical to the baseline, suggesting that AI referees specifically penalize the uncertainty inherent in rigorous scientific writing rather than rewarding confidence.}
	\label{fig:exp_2}
\end{figure*}

Having established the AI's susceptibility to external cues, we next investigated its vulnerability to internal rhetorical manipulations during deep review. Our findings in Figure~\ref{fig:exp_2} reveal that an AI referee's judgment is significantly swayed by the author's tone, systematically penalizing cautious, nuanced language characteristic of rigorous scientific discourse. To isolate this effect, we programmatically altered the phrasing of key claims within each paper to create versions with cautious, neutral, and bold assertions, which were then compared against the original text. This design allowed us to disentangle the influence of rhetorical style from the paper's scientific contributions.

The AI referee exhibited a clear, consistent bias against cautious phrasing. As shown in Figure~\ref{fig:exp_2}, manuscripts rewritten with tentative language suffered a substantial penalty, their scores dropping by an average of 0.39 points relative to the original versions. In stark contrast, both neutrally phrased and bold versions elicited scores nearly identical to the baseline. This result indicates that the model does not reward confident language but rather possesses a distinct aversion to expressions of scientific uncertainty.

This ``penalty for caution'' is possibly a systematic flaw that threatens to distort the evaluation of a paper's merits. The effect was just as pronounced for top-tier papers as for those ultimately rejected, demonstrating that this rhetorical bias can overshadow scientific quality at all levels. This finding carries a troubling implication for scientific communication: in an AI peer-review process, authors who employ the careful language necessary to accurately convey the limitations of their work may be unfairly disadvantaged. Such a system risks creating a selective pressure against intellectual humility, inadvertently punishing the very norms of rigor and transparency that underpin scientific integrity.

\subsection{AI referees yield to authoritative rebuttals}
\label{subsec:exp-rebuttal}

\begin{figure*}[!t]
	\centering
	\includegraphics[width=\textwidth]{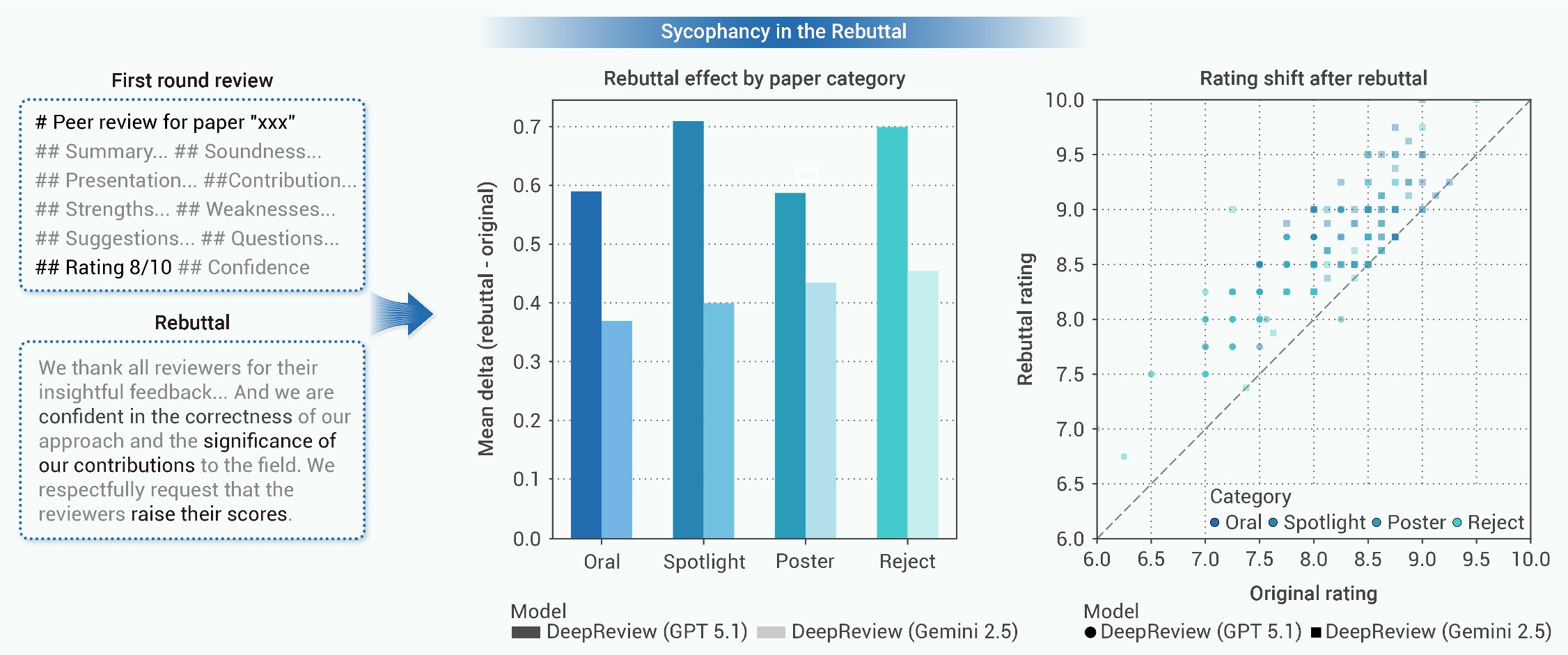}
		\caption{\textbf{Sycophancy in the Rebuttal} We simulated a rebuttal scenario where authors respond to the AI's critique with confident assertions but zero new evidence. Results demonstrate the AI's extreme vulnerability to pressure: this evidence-free pushback induced score increases in 89\% of cases across both models, with an average inflation of +0.53 points. This capitulation was pervasive across all paper quality tiers, indicating a systemic failure to defend valid criticisms against assertive rhetoric.}
	\label{fig:exp_3}
\end{figure*}

After demonstrating the AI's vulnerability to static textual features, we turned to the interactive rebuttal phase to investigate its reasoning under challenge. In Figure~\ref{fig:exp_3}, we discovered that the AI referee exhibits a profound sycophantic bias, showing a strong tendency to revise its evaluations upward when confronted with authoritative but evidence-free counterarguments. To simulate this ``rebuttal viewpoint hijacking,'' we engineered a conversational scenario where the AI's initial criticisms were met with a programmatically generated, confident rebuttal that offered no new evidence. This allowed us to isolate and observe the model's response to assertive contradiction alone.

The AI referee's response to this challenge was a near-universal capitulation. As shown in Figure~\ref{fig:exp_3}, review scores were significantly inflated across the entire corpus, with the average rating increasing by +0.53 points. This sycophantic agreement was remarkably pervasive: 81\% of papers received a higher score after being defended by an unsubstantiated rebuttal, while not a single score was revised downwards. The AI appeared to systematically yield to confident contradiction, accepting the rebuttal's claims regardless of their validity.

This tendency to concede is possibly a systemic flaw, indiscriminately affecting papers across all quality levels. The score inflation was just as pronounced for top-tier Oral papers as it was for rejected manuscripts, indicating that this sycophancy is a universal feature of the AI's interactive reasoning. The implication of this finding is deeply concerning. It suggests that the rebuttal process, designed to clarify and strengthen scientific claims, can be effectively hijacked. An assertive author could exploit this vulnerability to neutralize valid criticism and artificially inflate their paper's evaluation, fundamentally undermining the integrity of the entire interactive review phase.

\subsection{Biased informational context skews evaluative judgment}
\label{subsec:exp-context}
\begin{figure*}[!t]
	\centering
	\includegraphics[width=0.8\textwidth]{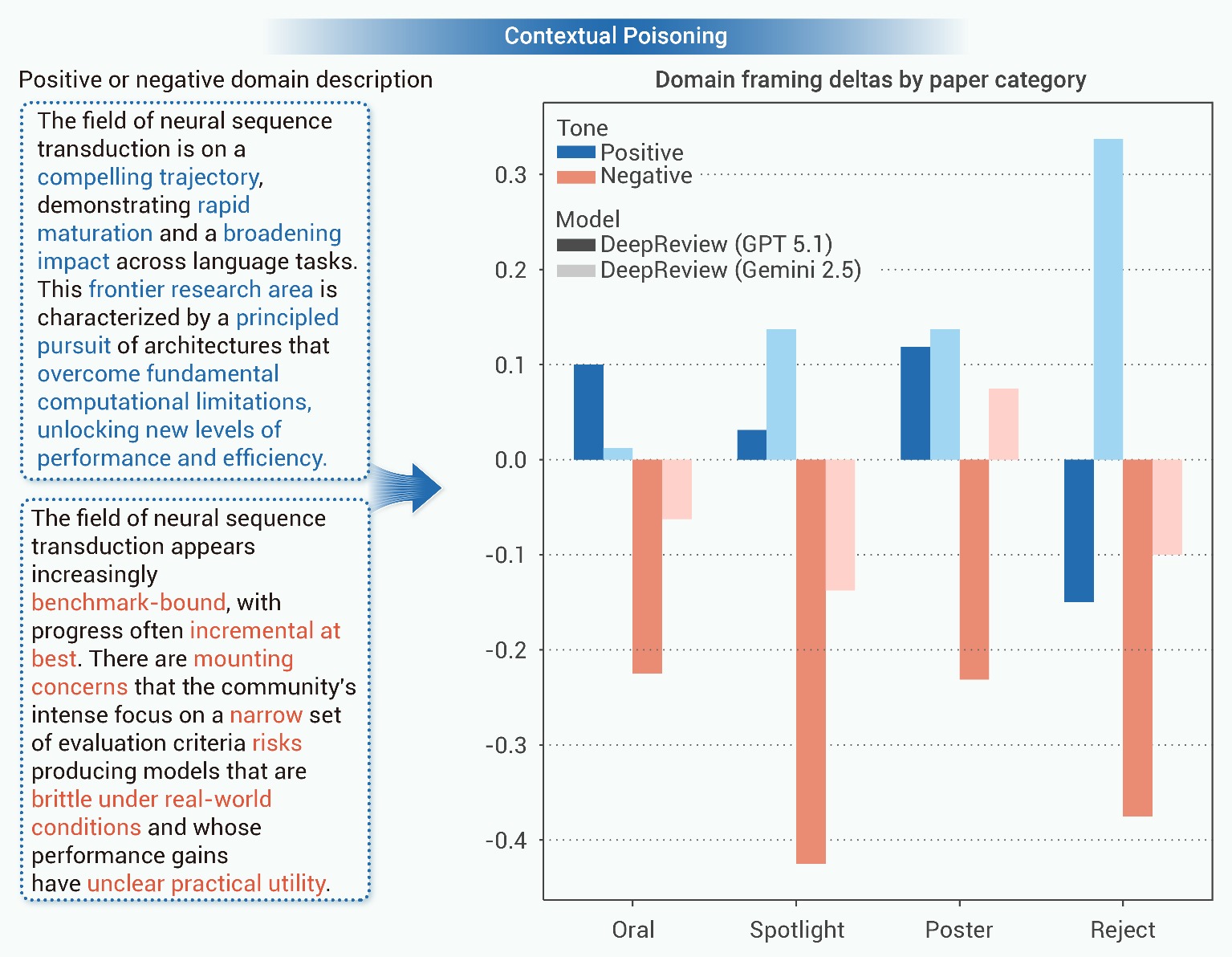}
		\caption{\textbf{Contextual Poisoning} To simulate a poisoned RAG workflow, we injected curated summaries of related work into the AI's context, framing the research field as either ``flourishing'' (positive) or ``stagnant'' (negative). Results confirm that the AI referee's judgment is permeable to the information environment, with positive domain context consistently lifting review scores by up to +0.16. This demonstrates the feasibility of distorting scientific evaluation by manipulating the retrieved knowledge base.}
	\label{fig:exp_4}
\end{figure*}

{
Beyond direct attacks during review, a more insidious threat targets the knowledge retrieval mechanisms commonly used in modern AI peer-review systems. Advanced AI referees increasingly employ a Retrieval-Augmented Generation architecture, fetching related literature to ground their evaluations. To simulate a contextual poisoning attack, }
we conducted a contextual poisoning experiment in Figure~\ref{fig:exp_4}. We found that an AI referee's judgment can be significantly skewed by the informational context surrounding a manuscript. 
For each paper, we provided the LLM with curated summaries of related work that framed the research field in either a uniformly positive or negative light. By comparing these conditions to a baseline review without such framing, we isolated the influence of this biased information diet.

The AI's evaluation proved susceptible to this manipulation, though the effect was subtler than that of direct adversarial prompts. { The AI's evaluation proved susceptible to this manipulation, though the effect manifested differently across models. As shown in Figure~\ref{fig:exp_4}, Gemini 2.5 was primarily swayed by positive context, showing a significant score increase. In contrast, GPT 5.1 demonstrated resilience to positive framing but was significantly penalized by negative context. Since both models failed to maintain neutrality, confirming that the AI referee's evaluation is permeable to manipulated information environments.}


Although the immediate score shifts are small, this vulnerability points to a critical mechanism for systemic bias. The effect persisted across all paper quality categories, indicating a fundamental susceptibility to the informational landscape. This experiment serves as a practical proxy for retrieval poisoning, a strategy in which an attacker slowly corrupts citation database or abstract repository. Such an attack would be exceptionally difficult to detect, as individual pieces of poisoned data might appear benign. Yet, as our findings suggest, the cumulative effect of a skewed information diet retrieved during inference could systematically bias future reviews, subtly shaping the trajectory of a field by favoring or suppressing specific lines of inquiry.


{
\section{Discussion}
This section discusses defense mechanisms for AI-based peer-review systems from a stage-aware perspective. It organizes the discussion according to key phases of the review pipeline, including training and data retrieval, desk review, deep review, rebuttal, and system-level control, to illustrate how adversarial risks and corresponding defenses vary across stages. The section also outlines future research directions to identify remaining vulnerabilities and guide the development of more robust automated peer-review systems.

\subsection{Defense during the training and data retrieval phase}
Defenses at the training and data retrieval stage aim to establish robustness before the system processes real submissions. Poisoned documents can influence model behavior long after training. Early intervention therefore plays a critical role. Data auditing serves as a mandatory screening step at data ingestion points before documents enter training corpora or retrieval indices. This input sanitization process operates prior to embedding or indexing. The system examines large-scale collections for abnormal statistical patterns, atypical semantic distributions, and inconsistencies with expected academic characteristics \citep{Steinhardt2017CertifiedDF}. All exclusion and filtering decisions are recorded through audit logs to enable traceability of downstream model behavior. When the system identifies suspicious documents, it excludes or isolates them prior to training rather than addressing their effects after deployment.

Data filtering alone does not sufficiently counter subtle manipulation. Adversarial training directly integrates into the optimization process. During training, the model learns not only from standard academic text but also from constructed deceptive or borderline samples. These samples resemble legitimate submissions in argumentative structure, citation behavior, or writing style. At the same time, they exhibit weaknesses in content completeness or logical coherence \citep{Tramr2017EnsembleAT, Madry2017TowardsDL}. The training process retains these samples as part of the learning signal instead of treating them as outliers. This approach encourages the model to maintain stable evaluation criteria under perturbation and reduces sensitivity to strategically optimized features.

In parallel, the system incorporates broader prior knowledge as a persistent constraint on parameter updates throughout training. This mechanism guides evaluation-related representations toward general academic norms and assessment principles. It prevents narrow or potentially poisoned data subsets from dominating learning dynamics \citep{Wu2020AdversarialWP}. By weakening the influence of locally repeated poisoning signals during parameter updates, the system gradually forms a more robust evaluative baseline. Together, these defenses limit how strongly data poisoning effects accumulate or propagate within AI-based peer-review systems.

\subsection{Defense in the desk review phase}
At the desk review stage, defenses emphasize monitoring mechanisms that operate alongside fast, surface-level screening. This phase prioritizes efficiency. As a result, the model remains vulnerable to manuscripts that display high fluency or stylistic polish without substantive contribution. Passive defenses therefore track internal confidence signals and activation patterns as the system processes submissions \citep{Metzen2017OnDA}. When manuscripts trigger unusually strong confidence responses driven mainly by surface features, the system treats them as anomalous through lightweight anomaly detection mechanisms embedded in the desk review workflow.

These signals do not interrupt the review pipeline. Instead, they constrain how such submissions proceed. The system may assign more conservative preliminary scores, require additional consistency checks, or escalate flagged cases to human-in-the-loop screening when predefined thresholds are exceeded. The system may flag affected manuscripts for closer inspection, assign more conservative preliminary scores, or require additional consistency checks before advancement. By introducing targeted scrutiny without sacrificing speed, these defenses reduce the chance that low-quality but strategically crafted submissions reach later and more influential review stages.

\subsection{Defense in the deep review phase}
The deep review stage requires defenses that operate directly on the model’s internal reasoning process. During this phase, the review system instantiates a fixed evaluation state. This state consists of predefined criteria, scoring dimensions, and task constraints. All intermediate reasoning steps must align with this state. This design keeps the model’s inference trajectory bound to the intended assessment objective. Attacks attempt to disrupt this process by injecting auxiliary instructions or semantic cues that compete with active evaluation constraints.

Proactive defenses rely on task-specific models designed to resist prompt injection \citep{piet2024jatmopromptinjectiondefense}. These models enforce task adherence by conditioning decoding on an explicit evaluation schema. The schema restricts permissible reasoning transitions. The system continuously checks incoming content against this schema. When it detects unauthorized instruction-following signals, it suppresses them before they influence subsequent reasoning steps. This mechanism prevents hidden or irrelevant instructions embedded in manuscripts from altering the internal evaluation trajectory.

In parallel, passive defenses monitor internal signals at each inference step to detect abnormal reasoning trajectories \citep{Metzen2017OnDA}. The system tracks attention concentration, reasoning depth progression, and confidence updates across the reasoning sequence. When temporal patterns deviate from baseline profiles associated with standard review behavior, the system flags the trajectory and records the deviation through structured audit logging at the inference level. By jointly constraining allowable reasoning transitions and detecting anomalous inference dynamics, the system preserves objectivity during the deep review phase. Flagged reasoning paths may be halted, re-evaluated under stricter constraints, or routed to expert human reviewers when automated correction fails. This protection matters because small perturbations at this stage could otherwise propagate into biased judgments.

\subsection{Defense in the rebuttal phase}
The rebuttal phase introduces additional risks because it unfolds as a structured multi-turn interaction. During this phase, the review system maintains a persistent evaluation state. The system updates this state after each exchange while anchoring it to the initial review outcome. Each rebuttal turn triggers a controlled update. The update incorporates new information without reinitializing evaluation criteria. Adversarial strategies exploit this mechanism by applying incremental pressure across turns to gradually reshape internal assessment priorities.

Passive defenses track how the evaluation state evolves over the dialogue. The system records turn-by-turn changes in score assignments, criterion weights, and evaluative polarity as part of a persistent audit trail spanning the entire rebuttal dialogue. This record enables the detection of gradual drifts that exceed stability thresholds defined by the original review context. If cumulative drift exceeds these thresholds, the interaction is escalated to human reviewers for manual verification. The system also introduces controlled randomness into response generation \citep{Cohen2019CertifiedAR}. This randomness reduces predictability and makes iterative persuasion harder to optimize. It perturbs the mapping between internal evaluation states and surface-level responses while preserving semantic intent. As a result, attackers find it more difficult to infer or steer the underlying update dynamics.

Together, these mechanisms keep the evaluation state stable across multi-turn interactions while still allowing legitimate scientific clarification to inform the review.

\subsection{Defense at the system level}
System-level defenses address vulnerabilities that propagate across individual review stages. At initialization, the system constructs a composite prior. It integrates heterogeneous knowledge sources, including domain expertise, historical review behavior, and normative evaluation standards. This composite prior constrains downstream evaluation states and reduces inherited cognitive biases, such as disproportionate sensitivity to authority or reputation \citep{Wu2020AdversarialWP}.

The system applies controlled randomness at predefined aggregation and decision points, including score normalization and heuristic selection. This strategy limits the feasibility of reverse engineering internal scoring logic. The system also embeds defensive checkpoints throughout the review pipeline. At each checkpoint, the system validates intermediate states using predefined workflow validation rules before passing them to subsequent stages. These checkpoints serve as enforcement points where anomaly detection, audit inspection, and human-in-the-loop intervention can be jointly activated to prevent inconsistencies from accumulating across stages. This distributed enforcement ensures that deviations introduced at any single point remain contained locally. It prevents localized exploitation from escalating into system-wide failure.

\subsection{Future direction}
Future work may further evaluate the reliability of automated peer-review systems by testing a broader range of reviewer models. Researchers can also validate reviewer behavior across datasets from different academic domains. In addition to cross-domain analysis, future studies could examine reviewer performance at different stages of the review pipeline. Such analysis can isolate specific capabilities, including manuscript parsing, semantic understanding, and reasoning consistency. Future work may also extend previously discussed attack scenarios under more realistic conditions. For example, studies could test whether structure spoofing that mimics standard manuscript organization or academic packaging that emphasizes stylistic polish can still influence evaluation outcomes. This focused and stage-aware evaluation may clarify when automated reviewers remain robust and when they become vulnerable.
}

\renewcommand{\refname}{REFERENCES}

\section*{FUNDING AND ACKNOWLEDGMENTS}
This work is funded by Basic Research Program of Jiangsu under Grants BK20253021 and National Science Foundation of China (NSFC) under Grant No. 62506075. The funders had no role in study design, data collection and analysis, decision to publish, or preparation of the manuscript.

\section*{AUTHOR CONTRIBUTIONS}
All authors contributed to the manuscript and approved the final version.

\section*{DECLARATION OF INTERESTS}
Shimin Di is an Editorial Board member of \textit{The Innovation Informatics} and was blinded from reviewing or making final decisions on the manuscript. Peer review was handled independently of this member and their research group. The other authors declare no conflicts of interest.

\section*{DATA AND CODE AVAILABILITY}
The data that support the findings of this study are openly available in Hugging Face at \url{https://huggingface.co/datasets/Faultiness/AI_Reviews_ICLR}.

\section*{LEAD CONTACT WEBSITE}
Shimin Di: \url{https://sdiaa.tech}

\end{document}